\documentclass[10pt,twocolumn,letterpaper]{article}

\usepackage{cvpr}
\usepackage{times}
\usepackage{epsfig}
\usepackage{graphicx}
\usepackage{amsmath}
\usepackage{amssymb}
\usepackage{multirow}
\usepackage{scrextend}
\usepackage[inline]{enumitem}
\usepackage{xcolor}
\usepackage[bold]{hhtensor}
\usepackage[T1]{fontenc}
\usepackage{color}
\usepackage{float}
\usepackage{placeins}
\graphicspath{{figures/}}
\usepackage[font=normal]{subfig}
\usepackage{enumitem}
\usepackage{authblk}

\newcommand*{\affaddr}[1]{#1} 
\newcommand*{\affmark}[1][*]{\textsuperscript{#1}}

\setenumerate[1]{itemsep=1.5pt,partopsep=0pt,parsep=\parskip,topsep=5pt}
\setitemize[1]{itemsep=1.5pt,partopsep=0pt,parsep=\parskip,topsep=5pt}
\setdescription{itemsep=0pt,partopsep=0pt,parsep=\parskip,topsep=5pt}

\usepackage[pagebackref=true,breaklinks=true,letterpaper=true,colorlinks,bookmarks=false]{hyperref}

\usepackage[breaklinks=true,bookmarks=false]{hyperref}

\cvprfinalcopy 

\def\cvprPaperID{****} 

\begin{document}

\title{When Semi-Supervised Learning Meets Transfer Learning: Training Strategies, Models and Datasets}

\author{%
	Hong-Yu Zhou\affmark[1]\quad Avital Oliver\affmark[2]\quad Jianxin Wu\affmark[3]\quad Yefeng Zheng\affmark[1] \vspace*{-0.1in}\\
	\affaddr{\affmark[1]YouTu Lab, Tencent}\quad
	\affaddr{\affmark[2]Google Brain}\quad
	\affaddr{\affmark[3]Nanjing University} \vspace*{0.05in}\\
	\tt\small{\{whuzhouhongyu, wujx2001, yefeng.zheng\}@gmail.com, avitalo@google.com}\\
}

\maketitle

\begin{abstract}
	Semi-Supervised Learning (SSL) has been proved to be an effective way to leverage both labeled and unlabeled data at the same time. Recent semi-supervised approaches focus on deep neural networks and have achieved promising results on several benchmarks: CIFAR10, CIFAR100 and SVHN. However, most of their experiments are based on models trained from scratch instead of pre-trained models. On the other hand, transfer learning has demonstrated its value when the target domain has limited labeled data. Here comes the intuitive question: is it possible to incorporate SSL when fine-tuning a pre-trained model? We comprehensively study how SSL methods starting from pretrained models perform under varying conditions, including training strategies, architecture choice and datasets. From this study, we obtain several interesting and useful observations.
	
	While practitioners have had an intuitive understanding of these observations, we do a comprehensive emperical analysis and demonstrate that: (1) the gains from SSL techniques over a fully-supervised baseline are smaller when trained from a pre-trained model than when trained from random initialization, (2) when the domain of the source data used to train the pre-trained model differs significantly from the domain of the target task, the gains from SSL are significantly higher and (3) some SSL methods are able to advance fully-supervised baselines (like Pseudo-Label).

	 We hope our studies can deepen the understanding of SSL research and facilitate the process of developing more effective SSL methods to utilize pre-trained models. Code is available at \url{https://github.com/funnyzhou/ssl-vs-pretrained-models}.
\end{abstract}

\section{Introduction}
Deep Neural Networks have been found to be quite effective for solving problems in the domain of computer vision~\cite{imagenet2009,rcnn,vgg,inception2015,inception2016,resnet2016}. One main reason is that deep models can ``digest'' large-scale labeled dataset, which was quite difficult using previous approaches. However, building a large image dataset can be very tedious and costly. Moreover, there are cases that image labels need expert experience and special devices. For example, medical images should be labeled by experienced doctors using specific medical instruments. Even then, some of the produced labels can be unreliable.

As people always want better performance for free, the community started to focus on how to make use of unlabeled images which are cheap and plentiful. Semi-Supervised Learning (SSL) is born with the ambition to learn from both labeled and unlabeled data simultaneously\footnote{In this paper, we only focus on the applications of SSL in computer vision. Though SSL has shown its strength in other areas, we will leave the task of describing those to other papers.}. By introducing unlabeled data to the learning process, SSL is able to exploit the regularity hidden in the data. When combined with traditional supervised methods, SSL based approaches have shown their ability to enhance performance without importing noticeable supervision.

Another technique for improving on training only with labeled examples, transfer learning, has been widely employed in various settings, especially computer vision related tasks. Unlike SSL, transfer learning is good at tackling learning problems arised in different domains. Thanks to the popularity of vast image datasets, e.g., the ImageNet~\cite{imagenet2009} and Places~\cite{places2017} dataset, the source domain is often large enough to share similar representations with the target domain in both low-level and high-level features. Thus models pre-trained on these datasets often have good initialization, and which enables them to surpass models trained from scratch in various open problems and competitions~\cite{fcn2015,fasterrcnn2015,tsn2016,bilinearpool2015,maskrcnn2017}.

\begin{figure}
	\centering
	\subfloat[]{\includegraphics[width=0.35\columnwidth]{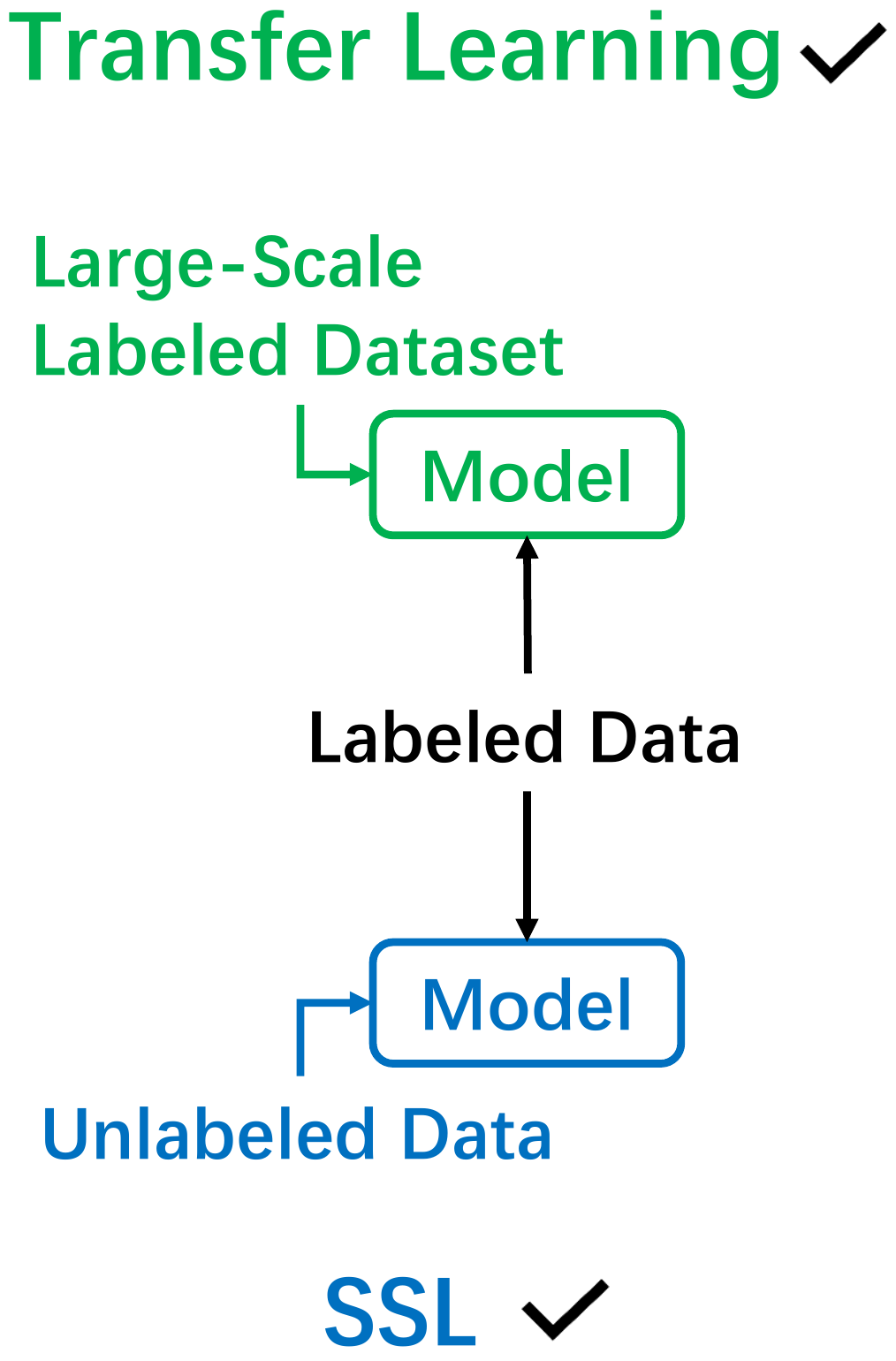}\label{left_intuition}}
	\quad
	\subfloat[]{\includegraphics[width=0.6\columnwidth]{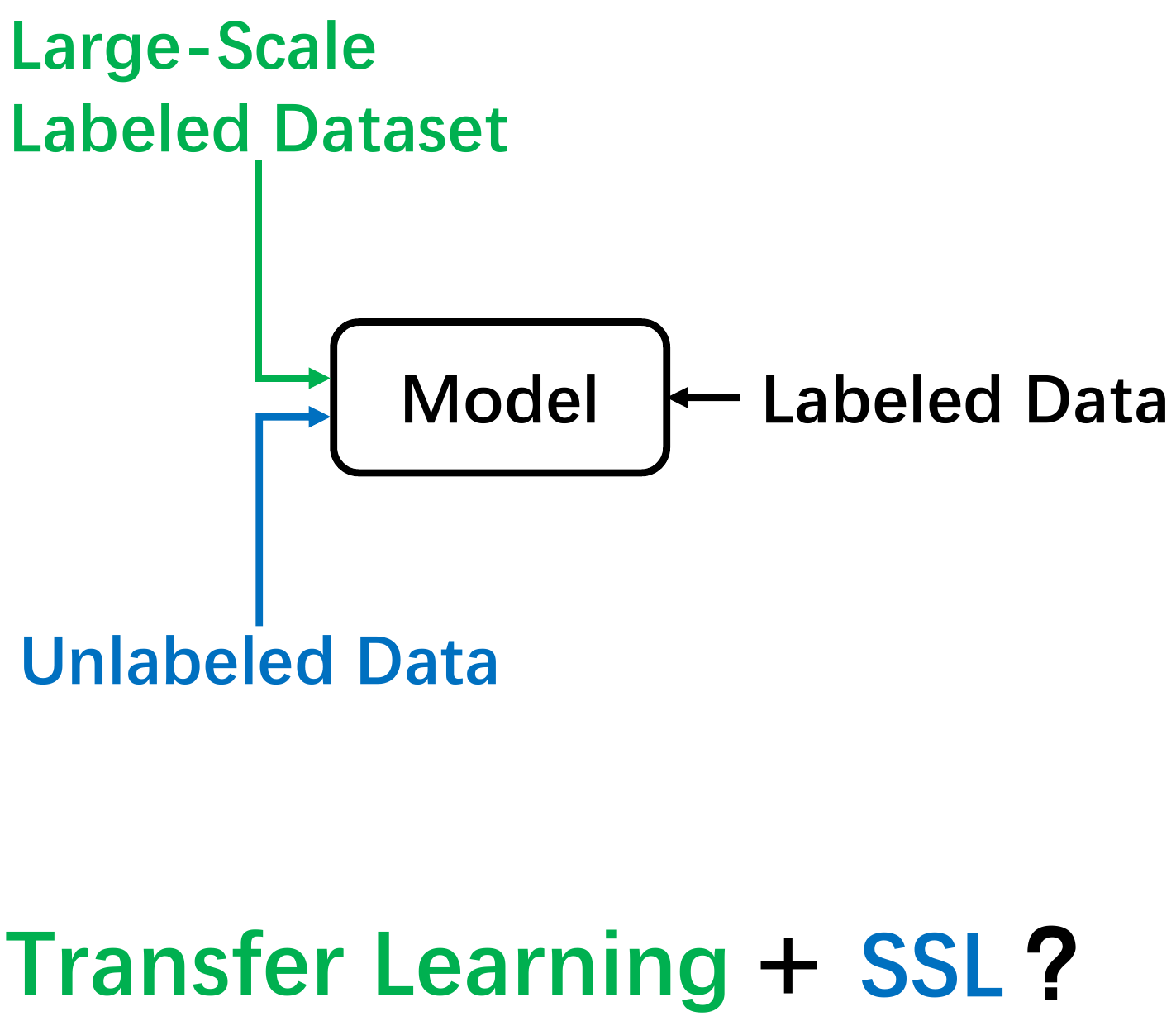}\label{right_intuition}}
	\caption{We present our core idea in this figure. From Figure~\ref{left_intuition}, we can see that the main difference between SSL and transfer learning is: SSL makes use of unlabeled images to facilitate the learning process in target domain, while transfer learning uses vast labeled dataset to learn generalizable representations in source domain. Since both two methods address learning better features for the target task, we want to check if they would have a conflict in real-world applications (as shown in Figure~\ref{right_intuition}).}
	\label{introduction}
\end{figure}

If we come to the topic of representation learning, it is not clear whether combining SSL and transfer learning would lead to learning better features. Then it is natural to wonder: would SSL benefit from transfer learning or would transfer learning from a large labeled dataset learn such a good classifier, such that SSL would offer little additional improvement? In fact, many recent works have demonstrated the effectiveness of SSL or transfer learning. But few studies tried to integrate both of them (as shown in Figure~\ref{introduction}). In this paper, we are trying to understand the effectiveness of SSL on top of transfer learning. We fine-tune pre-trained models using existing SSL approaches under varying training strategies, models and datasets. We report some interesting observations appeared in our experiments.

Our contributions can be summarized as follows:
\begin{itemize}
	\item We perform detailed experiments and discover that gains from modern SSL methods are smaller in settings where pre-training from a similar domain works well. Under many such experimental settings the gains entirely disappear.
	\item SSL-based algorithms work the best when the source domain is quite different from the target domain. Our experimental results show that SSL has great potential in processing medical images, where pre-training from ImageNet doesn't seem to lead to any significant improvements while SSL does.
	\item We find that Pseudo-Label with Adam as its optimizer, is surprisingly effective when you have enough labeled images. This fact gives the evidence that Pseudo-Label may be underestimated in previous literature. We caveat this point by pointing out that Oliver~\etal\cite{eval2018} find that Pseudo-Label worked poorly when training models from scratch, so additional experiments may be warranted here to tease apart the differences.
\end{itemize}

\section{Related Work}
\label{relatedwork}
Both semi-supervised learning and transfer learning are hot topics in the computer vision and machine learning communities. There are many classical approaches in these two areas, such as graph-based approaches~\cite{graphsemi2003} and transductive SVM~\cite{tsvm1999} in SSL, self-taught learning~\cite{selftaught2007} and graph transfer method~\cite{eigen2009} in transfer learning. Due to the lack of space, we only focus attention on widely adopted approaches with deep neural networks: $\Pi$-model~\cite{pi2016}, Mean Teacher~\cite{meanteacher2017}, Virtual Adversarial Training~\cite{vat2018}, Pseudo-Label~\cite{pseudo2013}, and fine-tuning.

\subsection{Semi-Supervised Learning}
Many methods have been proposed to tackle semi-supervised learning, we only introduce the most related ones. For example, we will not provide a comprehensive review of graph-based methods as they have not shows state-of-the-art results on recent SSL benchmarks with deep neural networks.
Recent studies in semi-supervised learning mainly lie in consistency-regularization and Generative Adversarial Networks (GANs). In this section, we also introduce entropy-based approaches and co-training.

\textbf{Consistency-Based Method}. Based on the network architecture presented in~\cite{valpola2015neural}, Rasmus~\etal\cite{ladder2015} proposed the ladder network, a model which is trained to simultaneously minimize the sum of supervised and unsupervised reconstruction cost functions by back-propagation. Laine and Aila~\etal\cite{pi2016} then simplified the ladder network to $\Pi$-model and introduced self-ensembling, a consensus prediction of the unknown labels using an exponential moving average of outputs of the network-in-training on different epochs. Tarvainen and Valpola~\cite{meanteacher2017} developed a method, named Mean Teacher, which keeps an exponential moving averages of model weights instead of the self-ensembling mentioned above.

Recently, Chen~\etal\cite{memory2018} proposed a method capable of exploiting the memory of a model and introduced a memory mechanism into the network training process.

\textbf{SSL using GANs and Adversarial Training}. The objective of GAN training is to generate visually realistic images, which seems to be a very suitable choice for SSL as these images can be taken as additional training data. Springenberg~\cite{firstsslgan2015} presented a method for learning a discriminative classifier from unlabeled or partially labeled data, which can be interpreted as the first attempt trying to apply GANs to SSL. Salimans~\etal\cite{ssgan2016} improved the techniques for training GANs and showed how a discriminator that also predicts classes can be used for SSL. Dai~\etal\cite{badgan2017} gave the definition of a preferred generator and derived a new formulation for improving previous feature matching GANs. Li~\etal\cite{triplegan2017} pointed out that a single discriminator only estimates the data without considering the labels and proposed Tri-GANs to address this problem. 

With respect to adversarial training, Miyato~\etal\cite{vat2018} modified the training course and proposed a new regularization method based on virtual adversarial loss. Park~\etal\cite{vatdropout2018} then introduced a minimal set of dropouts, called adversarial dropout, to further improve the performance of virtual adversarial training.

\textbf{Entropy-Based SSL}. Grandvalet and Bengio~\cite{emssl2005} considered entropy minimization as a regularizer to incorporate unlabeled data. Lee~\cite{pseudo2013} proposed a simple method, called Pseudo-Label, which just picks up high-confidence predictions which are iteratively added to the labeled training set.

\textbf{Co-Training}. Co-Training, first proposed by Blum and Mitchell~\cite{cotraining1998}, utilizes the diversity between two classifiers and let them label unlabeled data for each other. Zhou and Li~\cite{tritraining2005} presented Tri-Trainig to use bootstrap sampling
to get three different training sets and generates three classifiers from these three training sets respectively. For deep models, Chen~\cite{trinet2018} tried to build Tri-Net to combine tri-training with deep models. 

In this paper, we choose to evaluate SSL methods based on consistency-regularization, adversarial training and entropy-based methods. The reason why we ignore GANs series is that pre-trained models are \emph{not} widely used in such methodologies. To keep pace with~\cite{eval2018}, we mainly perform experiments on $\Pi$ model, Mean Teacher, VAT and Pseudo-Label.

\subsection{Transfer Learning}
According to~\cite{tfsurvey2010} and~\cite{tfsurvey2018}, research on transfer learning can be divided into different categories using different rules. Fine-tuning, which usually means applying the architecture and parameters of network pre-trained in source domain to the target domain, is recognized as the most commonly used technique under various datasets and network architectures. In this paper, we use fine-tuning as our main method of transfer learning.


Yosinski~\etal\cite{transfer2014} provided a thorough study about the fine-tuning performance across different network layers and varying image classes. As for the reason why we choose to fine-tune all layers, we argue that this fits the setting of SSL: training numerous parameters with the help from unlabeled data.

In the rest of this paper, we adhere to a similar idea proposed in~\cite{transfer2014} except that we incorporate SSL into the fine-tuning process. Our work shares some similarities with a recent evaluation paper on SSL~\cite{eval2018}, in which the authors made a comprehensive study about the performance of SSL on real-world applications. But their experiments were mostly based on models trained from scratch and reported few results about fine-tuning a pre-trained model under various conditions (e.g., different datasets and model architectures). In this paper, we expand their analysis to the combination of fine-tuning and SSL.

\section{Revisiting SSL Methods}
\label{sslrep}
Since there are so many SSL algorithms, it is necessary to select several representatives so as to perform further experiments. Oliver~\etal\cite{eval2018} chose two explicit consistency-regularization (or so called smooth regularization in \cite{badgan2017} and \cite{vat2018}) method: $\Pi$-model~\cite{pi2016} and Mean Teacher~\cite{meanteacher2017}, one consistency regularization approach based on adversarial perturbations: Virtual Adversarial Training (VAT)~\cite{vat2018} and the Pseudo-Label~\cite{pseudo2013} as experimental subjects. We think this algorithm pool sounds reasonable and adopt this setting in our studies. 

In this section, we explain how both VAT and Pseudo-Label may also be categorized as consistency-regularization SSL methods. We also argue that the regularization of SSL approaches can provide potential to further improve fully-supervised methods.

\subsection{Preliminary}
We use $D_{\text{L}}$ to denote a set of labeled images, and $D_{\text{UL}}$ to denote a set of unlabeled images. The original input image is $x_i$. To clarify the difference between labeled and unlabeled image, we use the following convention: The first elements in $D$ are the labeled images $\left\{x_{1,2,...,\text{n}}\right\} \in D_{\text{L}}$ and the later elements in $D$ are the unlabeled images $\left\{x_{\text{n+1,n+2,...,n+m}}\right\} \in D_{\text{UL}}$. The number of all images is $N$, which equals $n+m$.

In consistency-based SSL methods, we also have some transformed (or corrupted) images. We denote the transformed image as $\tilde{x}_i$ (since these transformations are stochastic, $x_i$ is a random variable). Note that $\tilde{x}_i$ comes from $x_i$ and hence $\tilde{x}_i$ can be written as $g(x_i, d_i)$, where $g(\cdot,\cdot)$ is a perturbation function and $d_i$ is a random variable sampled from some noise distribution.

When $d_i$ is independent of $x_i$, $g(x_i, d_i)$ can be written as $x_i*d_i$. For example, to describe random horizontal flips, let $d_i$ be sampled from Bernoulli distribution and $x_i * d_i$ be defined respectively as either $x_i$ or flipped $x_i$. 
We can extend our framework for dropout~\cite{dropout2014}, which is also used for consistency-based SSL approaches: Let $x_i^{(l)}$ be the input to layer $l$, and $d_i^{(l)}$ is sampled from a vector Bernoulli variables, one per neuron. Then $x_i^{(l)}*d_i^{(l)}$ is the layer when the selected neurons are dropped. However, for simplicity, we ignore dropout used in SSL methods in following equations.

The whole network can be described as producing an output class probability distribution $f_{\theta}(\cdot)$, where $\theta$ represents the network parameters, and we denote the number of classes by $k$.

We also use $L_{\text{CE}}[\cdot,\cdot]$ to represent cross-entropy loss, while $L_{\text{MSE}}[\cdot,\cdot]$ stands for the mean square error loss.

\subsection{$\Pi$ Models}
$\Pi$ model~\cite{pi2016} proposed adding an unsupervised loss term for the discrepancy between predictions of the model with different noise samples. The cost function of $\Pi$ model is:
\begin{align}
\label{pi}
\footnotesize
\begin{split}
L_{\Pi}(D_{\text{L}}, D_{\text{UL}}) = \sum_{i=1}^{N}L_{\text{MSE}}[f_{\theta}(x_i*d_i), f_{\theta}(x_i*\hat{d_i})] \\ + \quad
\alpha\sum_{i=1}^{n}L_{\text{CE}}[f_{\theta}(x_i*d_i), y_i]
\end{split}
\end{align}
In the formula above, we slightly abuse notation, where both $d_i$ and $\hat{d_i}$ refer to two samples from the same noise distribution.

\subsection{Mean Teacher Model}
The outputs of $\Gamma$ and $\Pi$ models are noisy during the initial training stage. Laine and Aila~\cite{pi2016} presented ``Temporal Ensembling'' to split $\Pi$ model into two different networks $f_{\theta}(\cdot)$ and $f_{\theta^{'}}(\cdot)$. This method accumulates outputs of each checkpoint (every epoch) and update predictions from $f_{\theta^{'}}(\cdot)$ by keeping exponential moving averages of the model parameter values. Tarvainen and Valpola~\cite{meanteacher2017} then made some modifications by simply substituting moving-average model weights for ensembled predictions. To incorporate such Mean Teacher method into formula~\eqref{pi}, we import a subscript $t$ into $\theta$, so the loss function becomes:
\begin{align}
\label{mt}
\footnotesize
\begin{split}
L_{\text{MT}}(D_{\text{L}}, D_{\text{UL}}) = \sum_{i=1}^{N}L_{\text{MSE}}[f_{\theta_{t}}(x_i*d_i), f_{\theta_{t}^{'}}(x_i*\hat{d_i})] \\ + \quad
\alpha\sum_{i=1}^{n}L_{\text{CE}}[f_{\theta_t}(x_i*d_i), y_i]
\end{split}
\end{align}
where $\theta^{'}_t=\beta\theta^{'}_{t-1}+(1-\beta)\theta_t$, $t$ stands for the training step and $\beta$ is a decay factor.

\subsection{VAT and Pseudo-Label}
VAT~\cite{vat2018} added a similar unlabeled loss term, with the following difference: instead of explicitly chosen perturbations, the perturbations are computed such that they maximally change the model's predictions. Hence, in VAT, the noise vector $d_i$ depends on $x_i$. And $L_{\text{VAT}}(\cdot,\cdot)$ is summarized as:
\begin{align}
\label{VAT}
\footnotesize
\begin{split}
L_{\text{VAT}}(D_{\text{L}}, D_{\text{UL}}) = \sum_{i=1}^{N}L_{\text{MSE}}[f_{\theta}(x_i), f_{\theta}(x_i * d_i)] \\ + \quad
\alpha\sum_{i=1}^{n}L_{\text{CE}}[f_{\theta}(x_i), y_i]
\end{split}
\end{align}
where $d_i$ is a sample from the conditional distribution $P(d|x_i)$ of virtual adversarial perturbations, computed through an approximation of the top eigenvalue of the Hessian of the KL of model predictions given input noise.

Pseudo-Label tries to select high-confidence predictions as the groundtruth labels and this process can be \emph{considered as a type of perturbation operation} on the output of $f_{\theta}(\cdot)$. While other perturbation make use of every unlabeled image, ``high-confidence'' only utilize a portion of unlabeled data. The loss criterion is as follows:
\begin{align}
\footnotesize
\label{PL}
\begin{split}
L_{\text{PL}}(D_{\text{L}}, D_{\text{UL}}) = \sum_{i=(n+1)}^{N}L_{\text{CE}}[f_{\theta}(x_i), g(f_{\theta}(x_i))] \\ + \quad
\alpha\sum_{i=1}^{n}L_{\text{CE}}[f_{\theta}(x_i), y_i]
\end{split}
\end{align}
where $g(\cdot)$ maps predicted class distribution $y$ into either $y$ (when $y$ is not confident), or a one-hot vector of the most confident prediction (when $y$ is confident past some specific threshold). 
By now, we are able to gather all these four methods into a consistency-regularization manner.

\section{Training Strategies, Models and Datasets}
Oliver~\etal\cite{eval2018} introduced a shared implementation of SSL algorithms and studied the performance of SSL under different data conditions, like distribution mismatch between labeled and unlabeled data, amount of labeled and unlabeled data, influence of having ImageNet for pre-training, and more. However, in this section, we extend the analysis to reveal more details about the combination of SSL and fine tuning from pre-trained models.

Our experiments mainly focus on three aspects: training strategies, models and datasets. These factors are at the core of machine learning experiments, and may lead to different conclusions 
from SSL experiments. This section will briefly outline these points and detailed analysis will be offered in the next section.

\subsection{Training Strategies}
We find that changes in training strategies may lead to huge differences in the insights gleaned from looking at the results of SSL training. In this section, we describe the factors that impacted
the intepretation of our results. More details will be reported in the following section.

\begin{enumerate}[label=(\alph*)]
	\item \label{optimzier} \textbf{Optimizer}. We made experiments on each setting with both Adam~\cite{adam2014} and SGD with Momentum.
	\item \label{labeleddata} \textbf{Amount of labeled data}. The number of labeled images is quite influential as reported in~\cite{eval2018}. We studied different numbers of labeled images across different datasets to check the influence those settings had on the final experimental results.
	\item \label{perturb} \textbf{Input Perturbation}. As mentioned in Section~\ref{sslrep}, manually designed perturbations are the basis of some of the consistency regularization SSL techniques we study, namely $\Pi$-model and Mean Teacher. We tested these SSL approaches under different perturbation methods to test how the experimental results depend on the perturbations used.
	\item \label{steps} \textbf{Training Iterations}. We find that SSL methods differ in how they perform when evaluated at different steps through the their training process. Some methods converge faster than others.
\end{enumerate}

We perform ablative experiments on these factors to show their influence on SSL. Note that we tune the baseline models and report results with appropriate learning rate and weight decay. More details can be found in Section~\ref{experiments}.

\subsection{Models}
Model architecture plays a significant role in the performance of SSL algorithms. For fairness, \cite{eval2018} used an identical 28-layer wide-resnet~\cite{wideresnet2016} to perform evaluation of different SSL algorithms. Nevertheless, we'd like to take a step forward: assessing different SSL algorithms under different network architectures.

\begin{table}
	\centering
	\caption{We report Top-1 and Top-5 Accuracy on ImageNet, floating point operations per second (FLOPs) and number of parameters in this table. Convolutional operations account for FLOPs. Inception-v3 has the highest Top-1 record while VGG-16 has the most parameters.}
	\label{modelsummary}
	\begin{tabular}{c|llll}
		Network & Top-1 & Top-5 & FLOPs & Params.\\
		\hline
		ResNet-50 & 75.2 & 92.2 & 8B & 26M \\
		VGG-16 & 71.5 & 89.8 & 32B & 138M \\
		Inception-v3 & 78.0 & 93.9 & 12B& 23M \\
	\end{tabular}
\end{table}
We give a brief summary of our candidates in Table~\ref{modelsummary}. ResNet-50~\cite{resnet2016} is a residual network where skip connects facilitate the training of deep models. VGG-16~\cite{vgg} is a popular architecture with fully-connected layers. Though it does not have high accuracy (at present), it is still active in many technique reports, which merits our attention. Inception-v3~\cite{inception2016} achieves the best performance over ResNet-50 and VGG-16 on ImageNet and was designed to be light-weight. We select these three models because they can be regarded as three important branches of the development of model architecture design. By testing on various model architectures, we better understand which techniques work well in particular settings, and which can be expected to improve results across the board.

\subsection{Datasets}
\begin{figure}
	\setlength{\belowcaptionskip}{-0.3cm}
	\centering
	{\includegraphics[width=0.8\columnwidth]{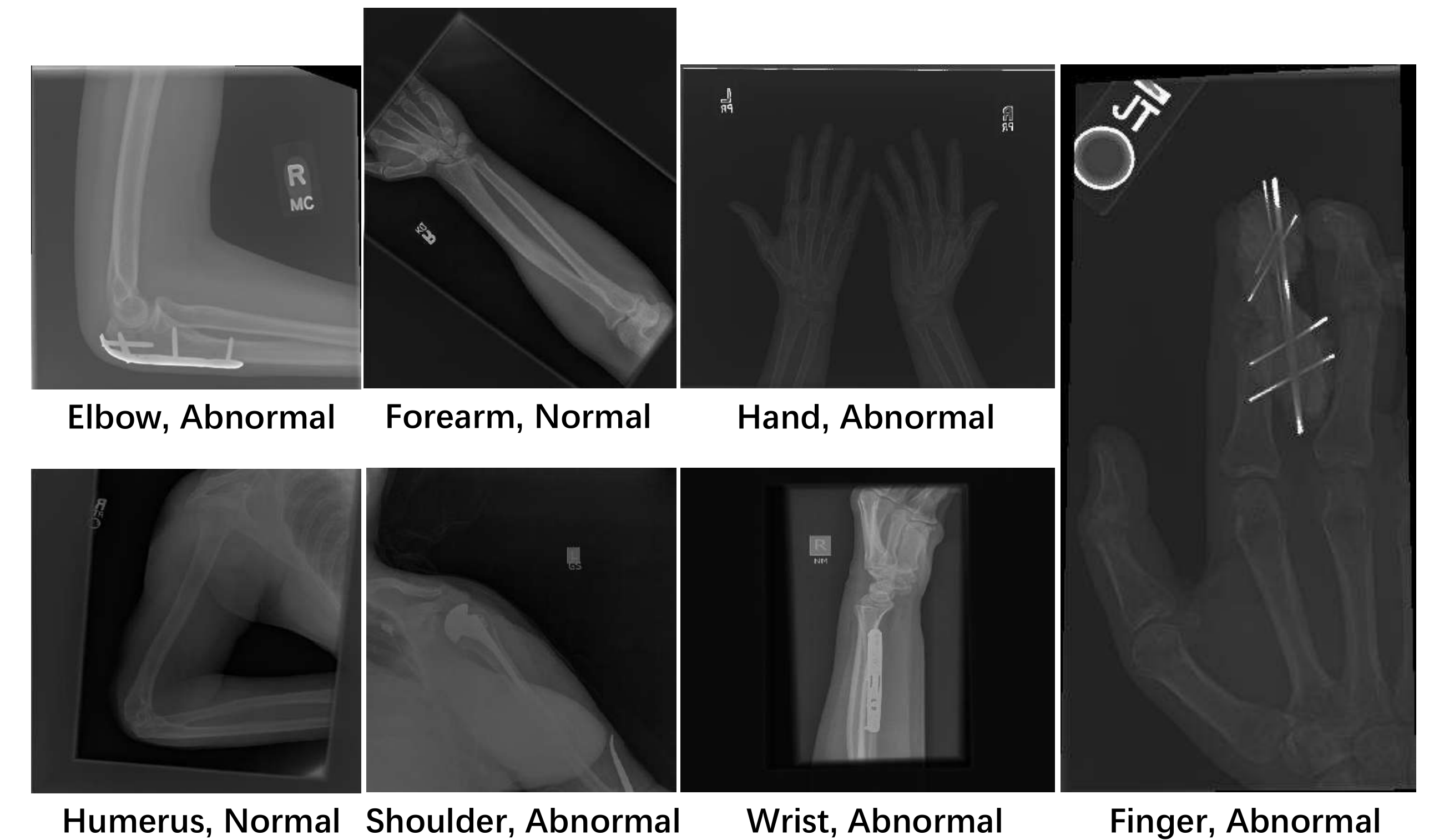}}
	\caption{Samples from MURA dataset. We show X-Rays of seven different parts from human body.}
	\label{sample_mura}
\end{figure}
As is commonly known, the gap between transfer learning and training from random initialization is highest when the target domain has a small number of labels. This overlaps with the setting where SSL is most helpful. Previous evaluation of SSL is done mostly on the MNIST, CIFAR and SVHN datasets. However, images in these datasets are usually small (less than 50$\times$50 in resolution), making them unsuitable for ImageNet-based models. In this paper, we use pre-trained ImageNet models and fine-tune on three modern datasets: Indoor67~\cite{indoor672009}, CUB200~\cite{cub2002011} and MURA~\cite{mura2017}. Note that Indoor67 and CUB200 have some class overlaps with ImageNet, while MURA is a medical image dataset which lies in a completely different domain.

\begin{figure*}
	\small
	\centering
	\subfloat[$\left\{\text{\_}/\text{25}/\text{1k}/\text{res50}/\text{indoor}/\text{F}\right\}$]{\includegraphics[width=0.6\columnwidth]{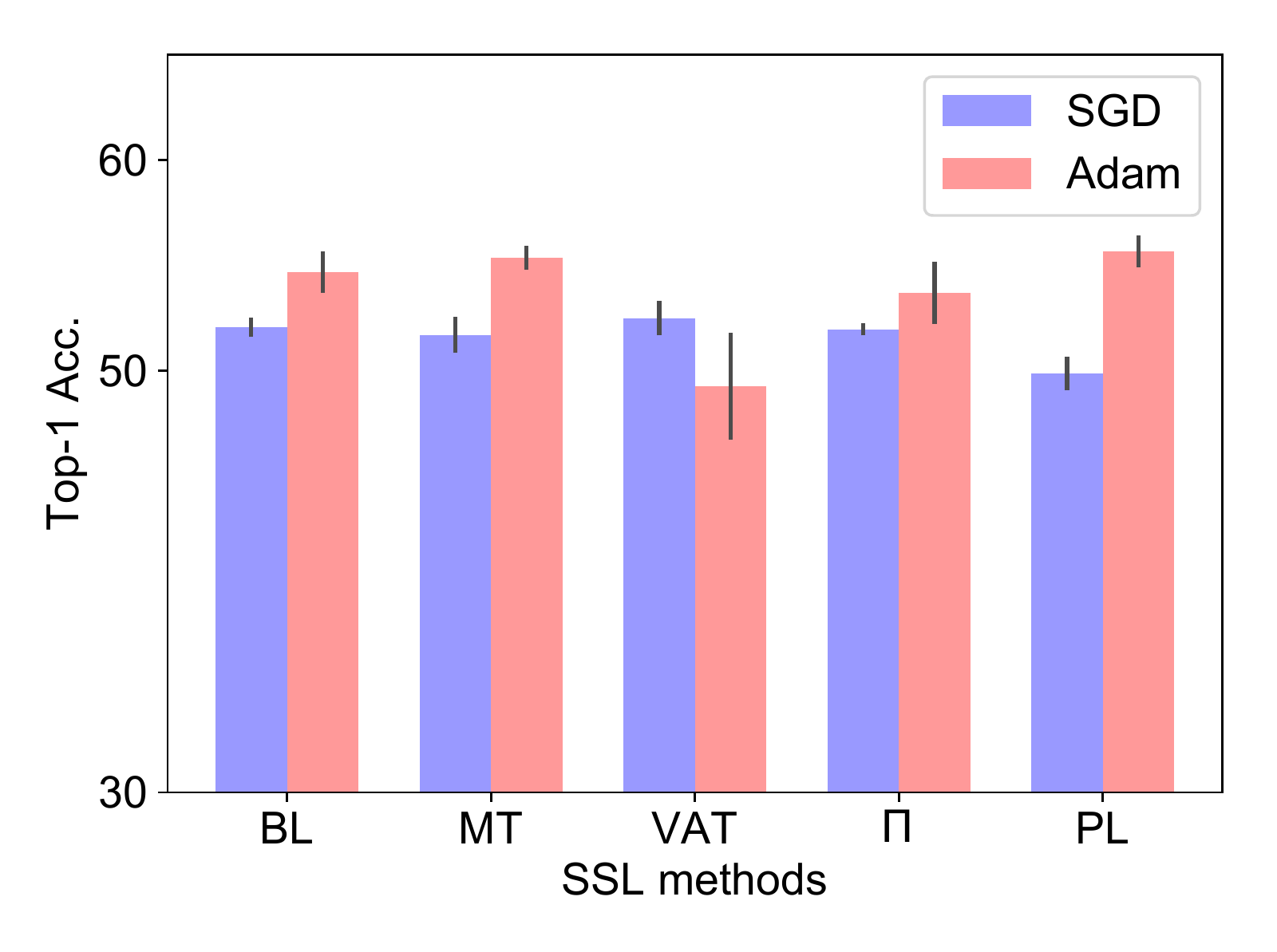}\label{opt_25_1k_res50_indoor_1}}
	\qquad
	\subfloat[$\left\{\text{\_}/\text{40}/\text{1k}/\text{res50}/\text{cub}/\text{F+T}\right\}$]{\includegraphics[width=0.6\columnwidth]{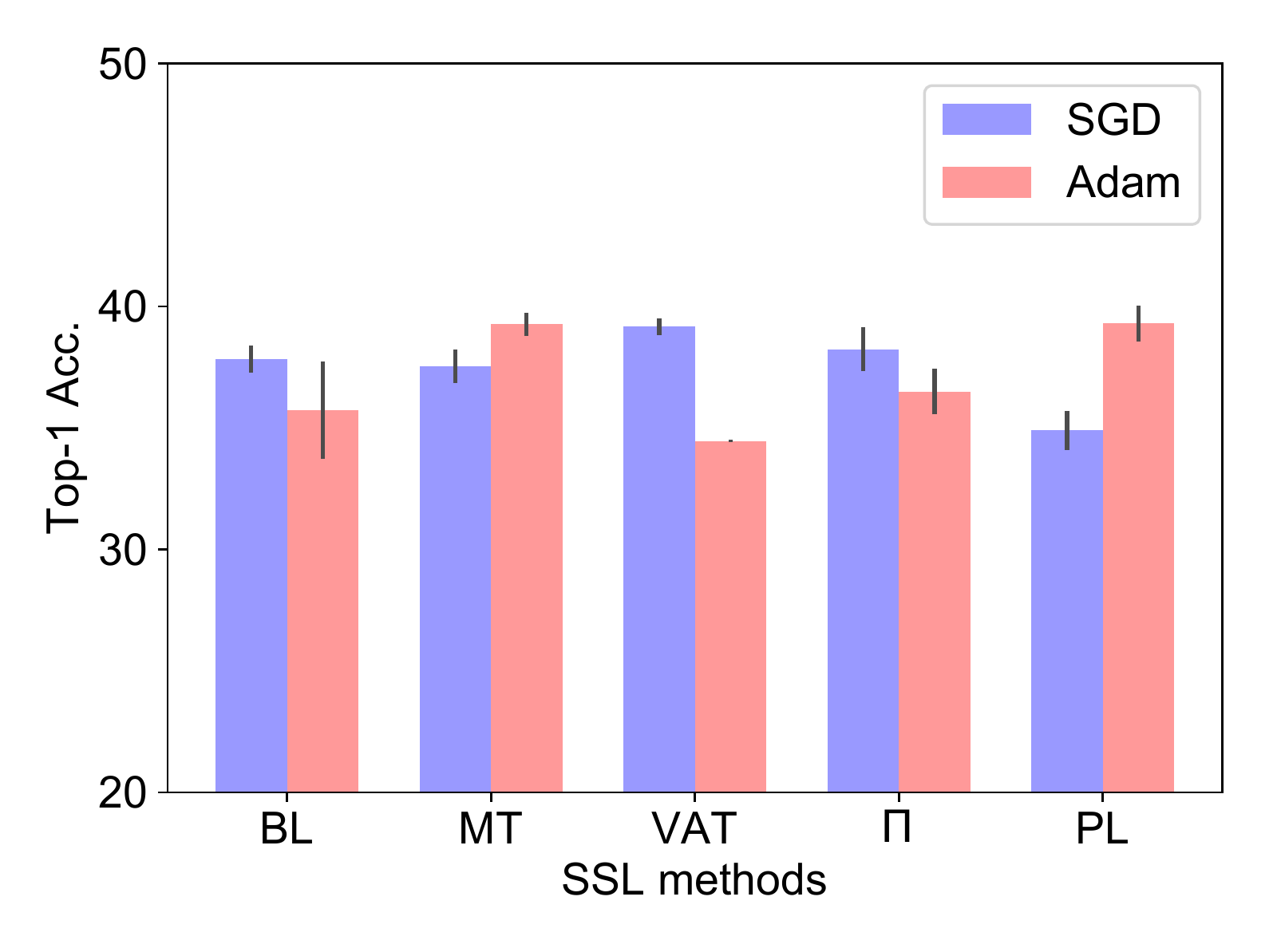}\label{opt_40_1k_res50_cub_2}}
	\qquad
	\subfloat[$\left\{\text{\_}/\text{40}/\text{1k}/\text{res50}/\text{mura}/\text{F+T}\right\}$]{\includegraphics[width=0.6\columnwidth]{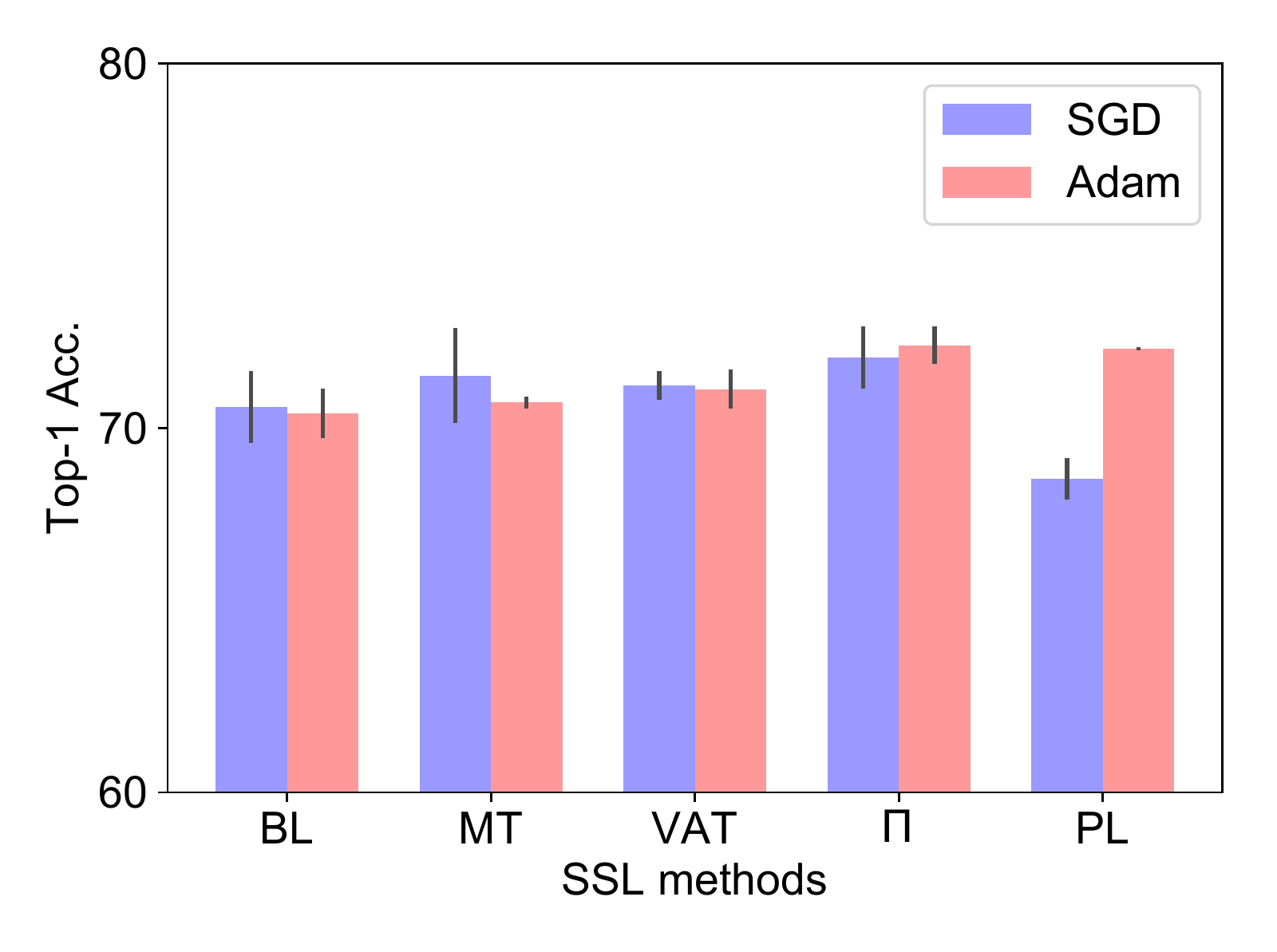}\label{opt_40_1k_res50_mura_2}} \\
	\subfloat[$\left\{\text{\_}/\text{40}/\text{3k}/\text{res50}/\text{indoor}/\text{F+T}\right\}$]{\includegraphics[width=0.6\columnwidth]{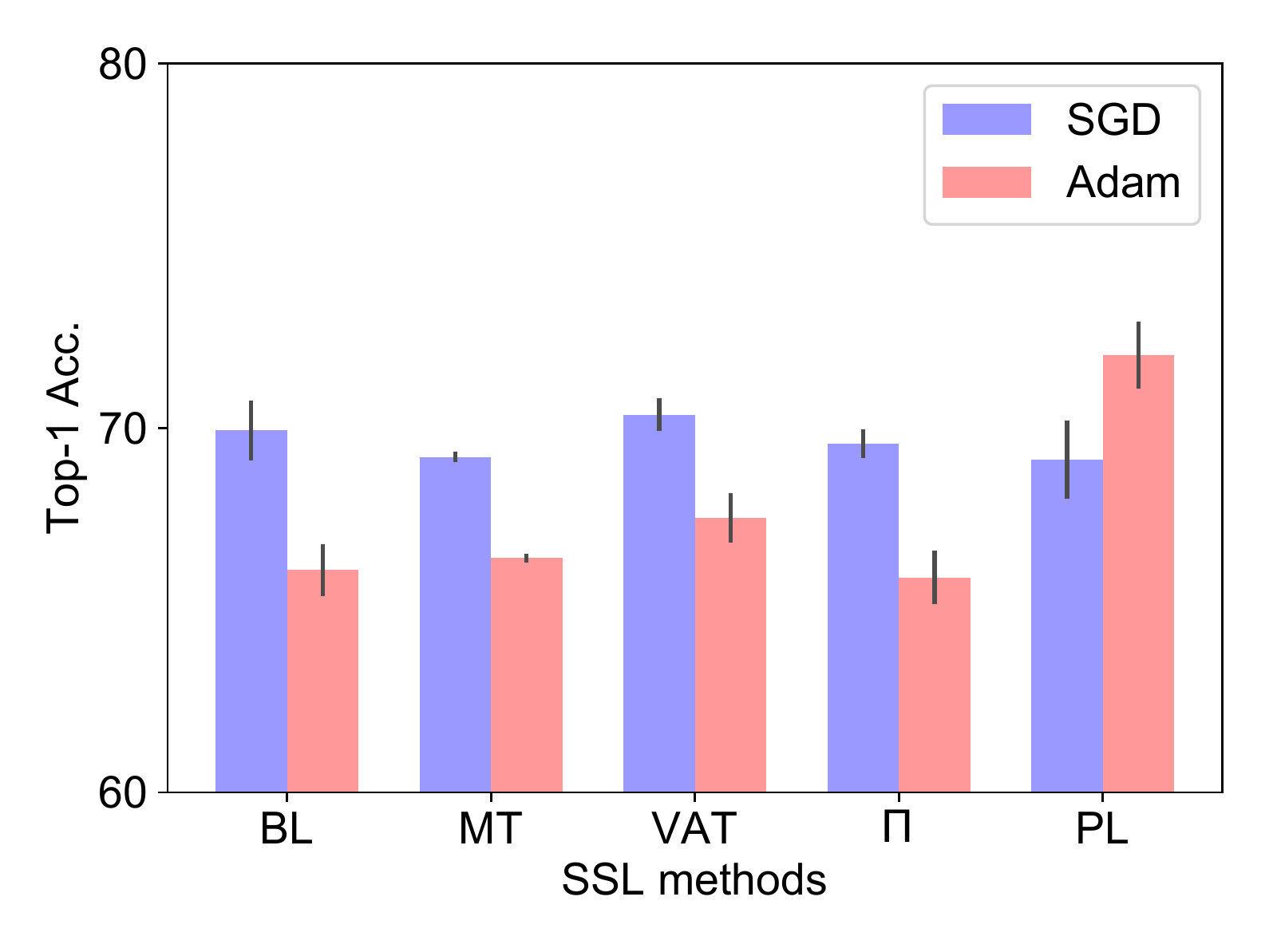}\label{opt_40_3k_res50_indoor_2}}
	\qquad
	\subfloat[$\left\{\text{\_}/\text{40}/\text{4k}/\text{res50}/\text{cub}/\text{F+T}\right\}$]{\includegraphics[width=0.6\columnwidth]{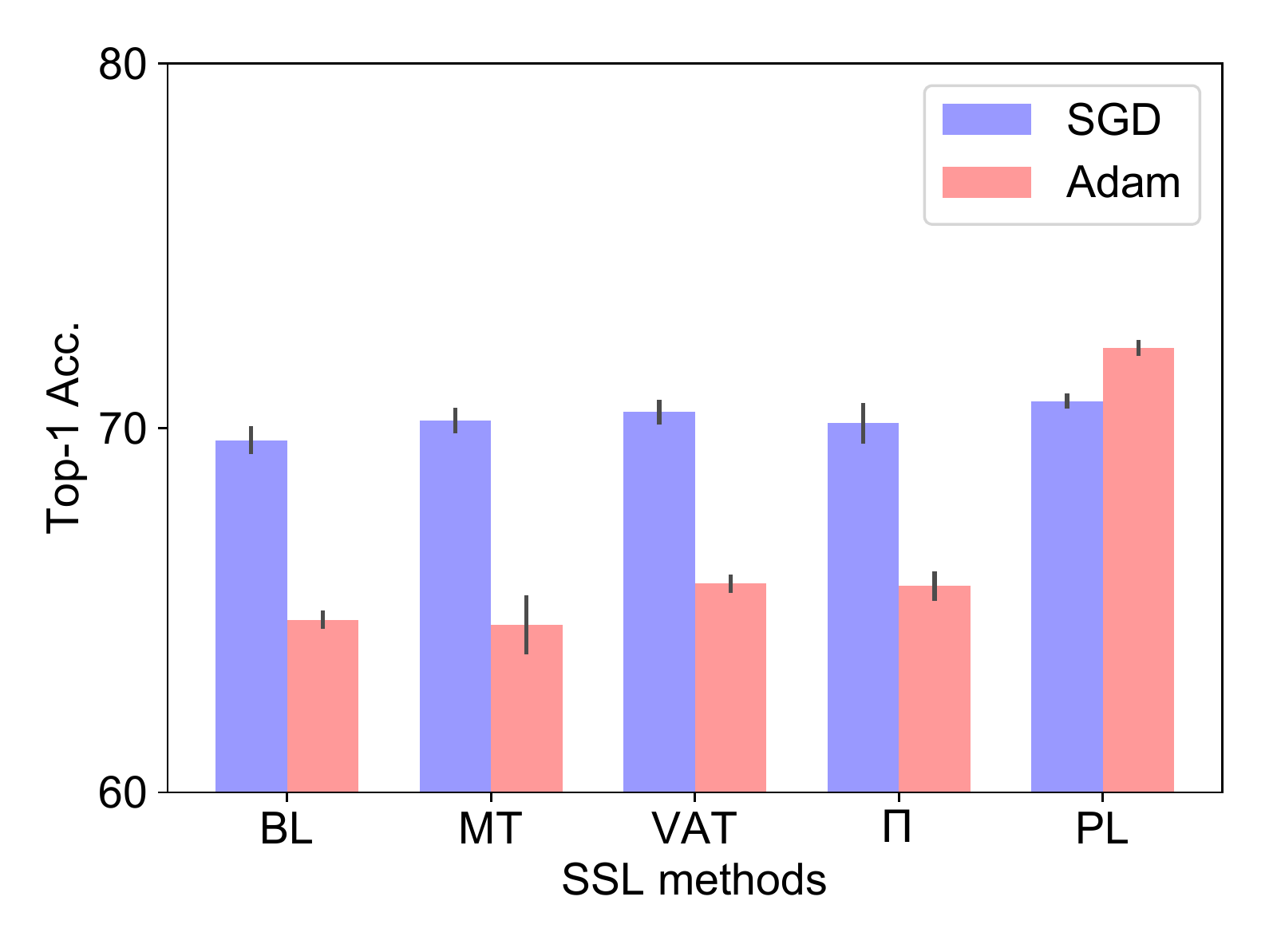}\label{opt_40_4k_res50_cub_2}}
	\qquad
	\subfloat[$\left\{\text{\_}/\text{40}/\text{2k}/\text{res50}/\text{mura}/\text{F+T}\right\}$]{\includegraphics[width=0.6\columnwidth]{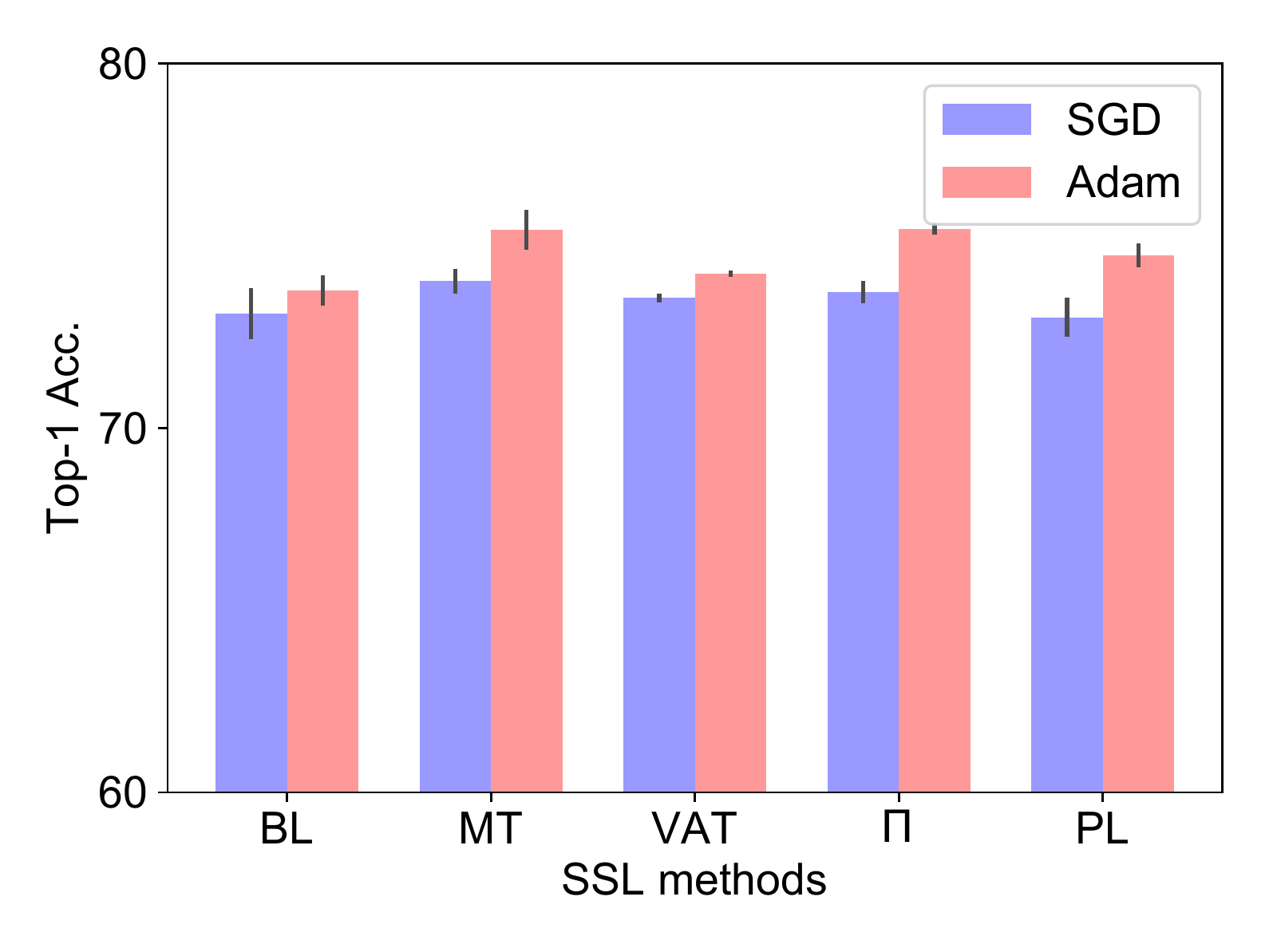}\label{opt_40_2k_res50_mura_2}}
	\caption{We performed experiments on different optimizers. Two conclusions are straightforward: 1. Pseudo-Label with Adam optimizer achieves the best results. 2. SGD surpasses Adam when increasing the number of labeled images and training iterations.}
	\label{adam_sgd}
\end{figure*}

Indoor67 is a scene-oriented dataset which is different from object-centered ImageNet. It contains 67 scene categories. There are 6,700 images in total, where each class has 80 images for training and 20 images for testing.

CUB200 is an object-oriented dataset with 200 species of natural birds. CUB200 includes 11,788 images and the official split for train/test is 5,994 vs. 5,794. Our experiments followed this convention.

MURA is a dataset of musculoskeletal radiographs, which contains 40,561 images from 14,863 patient studies. X-Ray images are collected from seven parts of human body: elbow, finger, forearm, hand, humerus, shoulder and wrist (cf. Figure~\ref{sample_mura}). The goal of this dataset is to distinguish normal musculoskeletal \emph{studies} from abnormal ones (a study often contains more than one image). In this paper, we make a modification to this goal: to simply tell the difference between normal and abnormal \emph{radiographs} (one image). The reason why we'd like to evaluate SSL on MURA is that we hope to check the value of SSL when ImageNet pre-trained models are not that useful.
\section{Experiments}
\label{experiments}
\begin{figure*}
	\small
	\centering
	\subfloat[$\left\{\text{SGD}/\text{40}/\text{\_}/\text{res50}/\text{indoor}/\text{F+T}\right\}$]{\includegraphics[width=0.6\columnwidth]{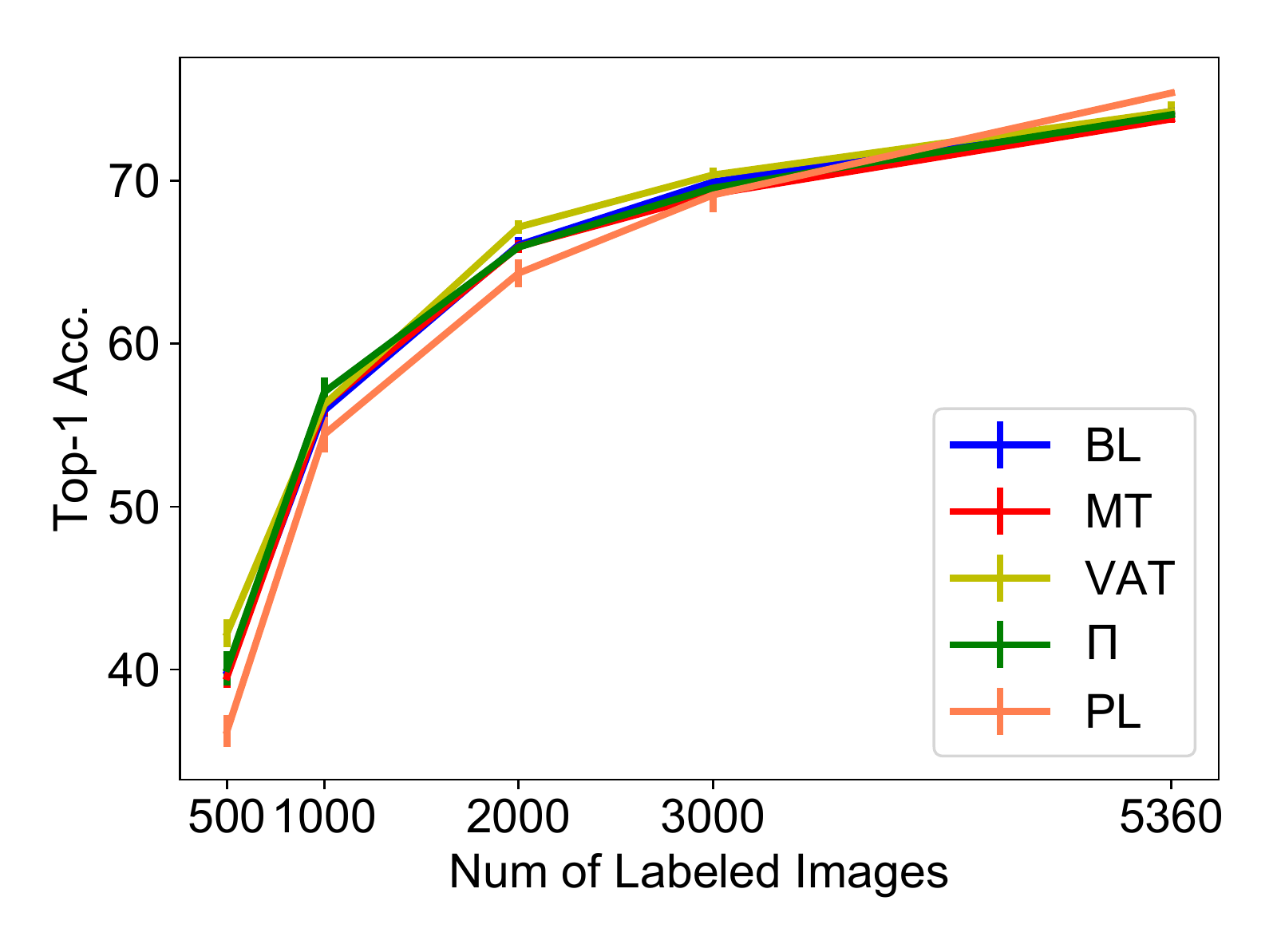}\label{sgd_40_amount_res50_indoor_2}}
	\qquad
	\subfloat[$\left\{\text{SGD}/\text{40}/\text{\_}/\text{res50}/\text{cub}/\text{F+T}\right\}$]{\includegraphics[width=0.6\columnwidth]{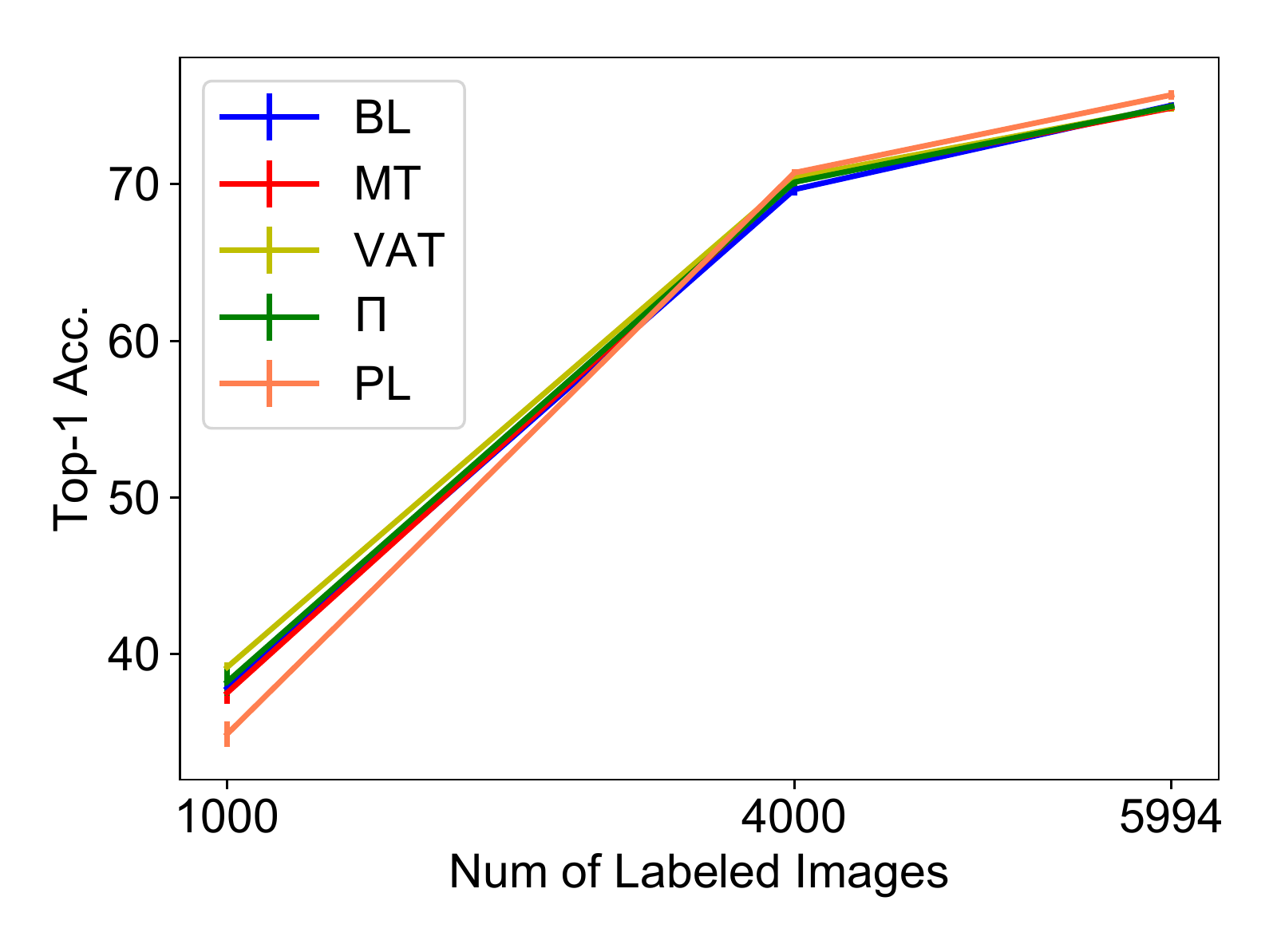}\label{sgd_40_amount_res50_cub_2}}
	\qquad
	\subfloat[$\left\{\text{SGD}/\text{40}/\text{\_}/\text{res50}/\text{mura}/\text{F+T}\right\}$]{\includegraphics[width=0.6\columnwidth]{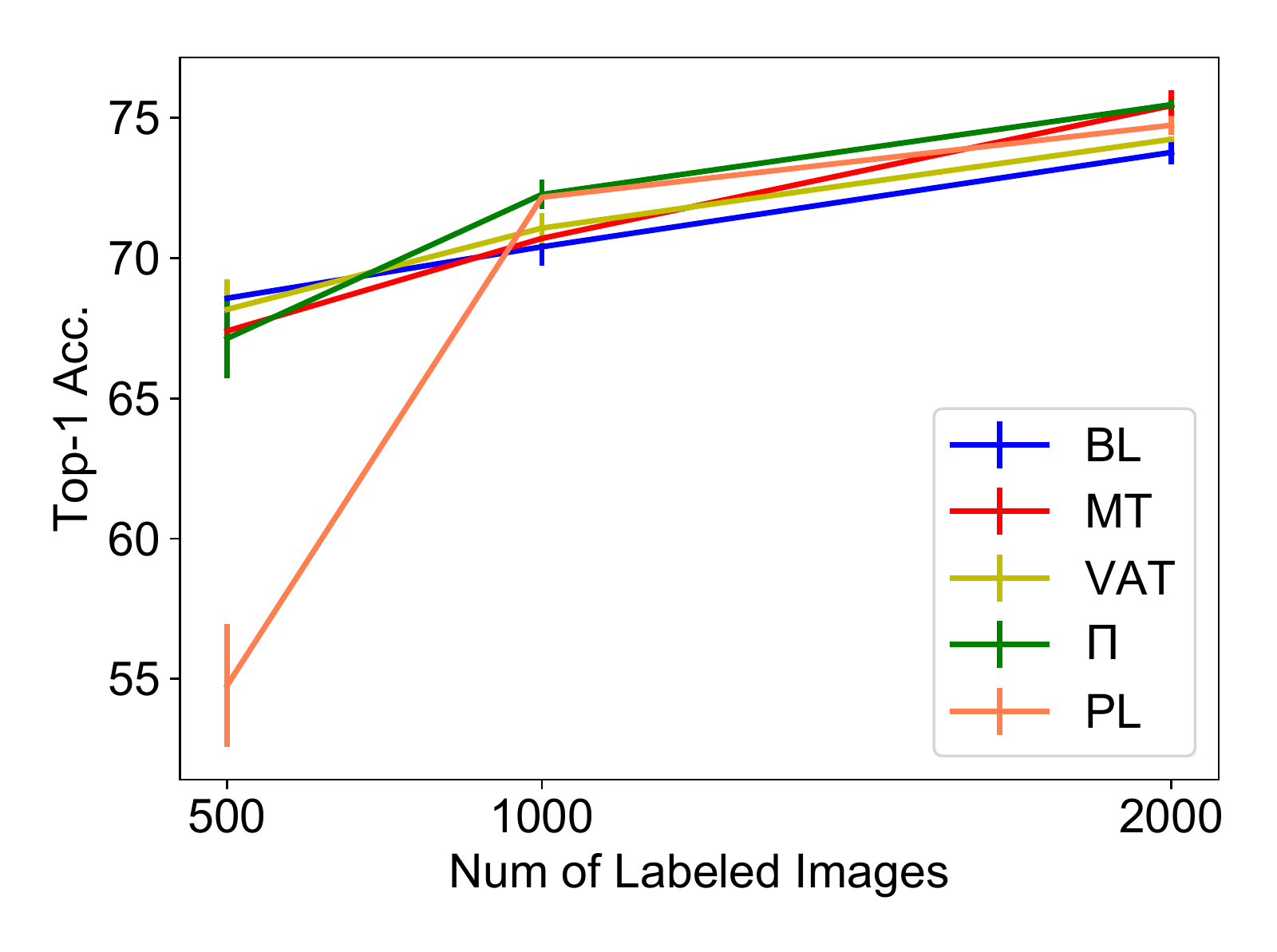}\label{sgd_40_amount_res50_mura_2}}
	\caption{Comparison between fully-supervised model and SSL algorithms with varying amount of labeled data.}
	\label{amount}
	
\end{figure*}
In this section, we mainly report the performance of SSL methods on pre-trained models. Before we start experiments, a few notations have to be introduced. For the abbreviation of SSL algorithms, we use $\Pi$ to stand for $\Pi$ model, \texttt{MT} for Mean Teacher, \texttt{VAT} for Virtual Adversarial Training, \texttt{PL} for Pseudo-Label and \texttt{BL} for fully-supervised method. In following experiments, we will make use of a set to denote hyperparameters during the training stage, 
\begin{equation*}
\label{set}
\left\{\text{Adam}/\text{20}/\text{1k}/\text{res50}/\text{indoor}/\text{F+T+N}\right\}
\end{equation*}

\begin{itemize}
	\item \texttt{Adam} means we employed Adam for optimization while \texttt{SGD} represents SGD optimizer.
	\item \texttt{20} denotes we trained both SSL methods and fully-supervised model for 20 epochs (similar strategy as in~\cite{eval2018}).
	\item \texttt{1k} means we have one thousand labeled images while the rest are unlabeled.
	\item \texttt{indoor} stands for the fine-tuned dataset. The other two databases are \texttt{cub} and \texttt{mura}, respectively.
	\item The last \texttt{F+T+N} tells us how many augmentation methods we have used. \texttt{F} means we only used horizontal flip, \texttt{T} indicates random translation, while \texttt{N} represents gaussian noise.
\end{itemize}

In summary, the above set means we trained all models (SSL and fully-supervised methods) based on Adam for 20 epochs using 1k labeled images. The pre-trained model is resnet-50 and we fine-tuned it on Indoor67 with horizontal flips, random translations, and Gaussian noise. At times, we refer to a class of experiments where only one setting varies -- we denote the varying setting with an underscore. We set the Mean Teacher exponential moving average decay rate to 0.99. The learning rates of Adam and SGD are 1e-4 and 1e-3, respectively -- we explain below how we standardized on these. Learning rates are decayed by a factor of 10 at three quarters of the total training steps. The size of input image is 224 on ResNet-50 and VGG-16, and 299 on Inception-v3. Other hyperparameters follow the settings in \cite{eval2018}.

The learning rates for Adam and SGD were chosen this way: We chose 1e-4 out of \{5e-3, 1e-3, 5e-4, 1e-4, 1e-5\} for Adam, and 1e-3 from \{5e-2, 1e-2, 5e-3, 1e-3, 1e-4\}. Out of those learning rates, we chose the one that performed best on $\{\text{\_/25/1k/res50/indoor/F\}}$ (see Figure~\ref{opt_25_1k_res50_indoor_1}). Then we fixed those learning rates for all other experiments. While we acknowledge that more careful tuning for each task may yield different results, we aimed to simulate the experience of practicioners that may not have the resources needed to tune for every new dataset. This methodology is also the one used in~\cite{eval2018}, where they fixed hyperparameters for a fixed number of labeled examples and evaluated with those same hyperparameters on other SSL settings.

The Adam hyperparameters are kept as in \cite{adam2014}: we set $\beta_1$ to 0.9 and $\beta_2$ to 0.999 in Adam.
The SGD momentum is kept at 0.9. Note that we don't tune the Momentum carefully as Wilson~\etal\cite{valueofoptimizer} showed that initial step size and the step decay scheme heavily influence the final performance.

\subsection{Adam versus SGD}
\label{adamvssgd}
We evaluated SSL approaches using different optimizers and display results in Figure~\ref{adam_sgd}. From this figure, we can see that without re-tuning for each setting, different SSL methods seem to work better with different optimizers. When comparing Figure~\ref{opt_25_1k_res50_indoor_1} with \ref{opt_40_3k_res50_indoor_2} and \ref{opt_40_1k_res50_cub_2} with \ref{opt_40_4k_res50_cub_2}, it is easy tell that \emph{simply increasing training iterations and amount of labeled data} may help SGD surpass Adam in all SSL approaches except Pseudo-Label. 

In fact, if we look carefully at the six figures, we see that \emph{Pseudo-Label with Adam} usually achieves the best results (sometimes quite close to the best) in all settings. This phenomenon is strange as Pseudo-Label got the worst performance on models trained from random initialization~\cite{eval2018}. We guess the reason might be that Pseudo-Label acts as a regularizer by adding a certain form of noise into the training process. This characteristic makes Pseudo-Label unable to achieve satisfying performance on models trained from scratch but helps prevent overfitting in pre-trained models. More details can be found in the appendix.

Another fact is that if we temporarily ignore Pseudo-Label, \emph{the other three SSL methods are comparable with simple baselines}. The ``best'' SSL candidate barely improves over the fully-supervised baseline. This phenomenon may complement the conclusion from \cite{eval2018} which tells us good pre-trained model could beat models trained from random initialization with SSL. However, our experiments demonstrate evidence towards the claim that {\textbf{under domain gap, you should not expect too much improvement from SSL on pre-trained models even with lots of unlabeled images}}.

When we turn to Figure~\ref{opt_40_2k_res50_mura_2}, we see that SSL offers more improvement when the target domain (radiograph) is different from the source domain (natural image). Nearly all SSL methods surpass the strong baseline where Mean Teacher and $\Pi$-model perform the best.

\subsection{The Number of Labeled Images}
In this part, we will use SGD as the default optimizer because SGD usually gets better accuracy when having more iterations and more labeled images as reported in Section~\ref{adamvssgd}. Figure~\ref{amount} shows cross-dataset performance when increasing the number of labeled images.

Still, the difference between the fully-supervised baseline and SSL is small. In Figure~\ref{sgd_40_amount_res50_indoor_2}, when the number of labeled images increases, \emph{Pseudo-Label gradually surpasses other methods including the baseline}. A similar phenomenon also appears in Figure~\ref{sgd_40_amount_res50_cub_2}. We argue that Pseudo-Label may not be good at making use of large amount of unlabeled data but succeed over other SSL methods when there are enough labeled images. For example, in Figure~\ref{sgd_40_amount_res50_indoor_2} and \ref{sgd_40_amount_res50_cub_2} (see the rightmost datapoints in the graph, that correspond to training on the entire dataset), Pseudo-Label even surpasses fully-supervised models with no labeled images in all three datasets. Since we borrowed the code of Pseudo-Label from~\cite{eval2018}, we argue the reason of this phenomenon is that they used pseudo-label in a way that adds two loss terms for all labeled examples, one based on the true label and one based on the confident predictions when they exist, while traditional Pseudo-Label separates the labeled and unlabeled data entirely. This implementation may provide a regularization effect on fully-supervised models.

In Figure~\ref{sgd_40_amount_res50_mura_2}, properties of Pseudo-Label become more clear. And we we can find that SSL methods surpass the fully-supervised baseline with different amounts of labeled data. This phenomenon shows us again that SSL improves the performance of pre-trained models when there is a domain gap.

\subsection{Models}
\begin{figure*}[htp]
	\scriptsize
	\centering
	\subfloat[\scriptsize $\left\{\text{SGD}/\text{40}/\text{1k}/\text{\_}/\text{indoor}/\text{F+T}\right\}$]{\includegraphics[width=0.49\columnwidth]{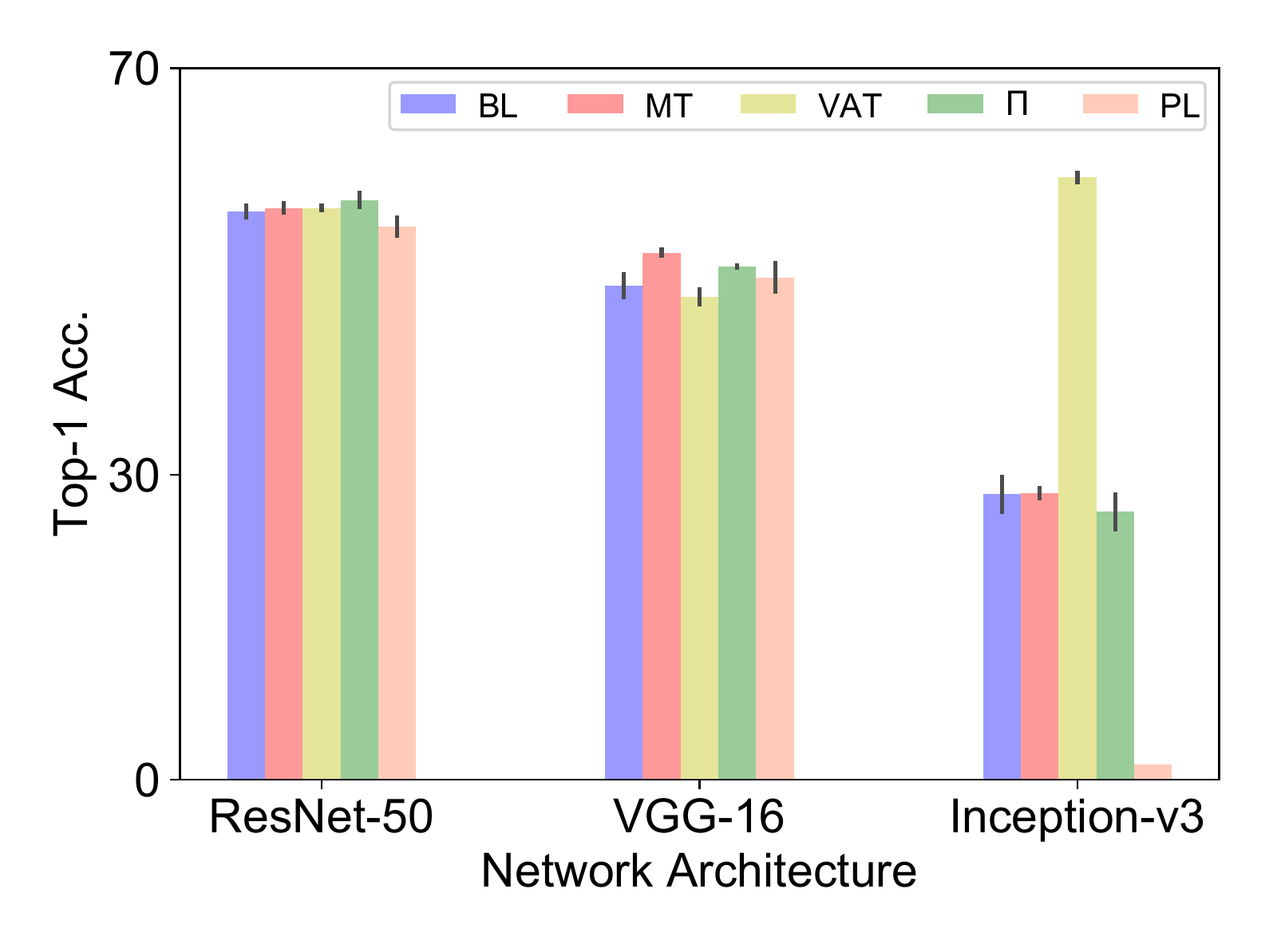}\label{sgd_40_1k_models_indoor_2}}
	\quad
	\subfloat[\scriptsize $\left\{\text{SGD}/\text{40}/\text{3k}/\text{\_}/\text{indoor}/\text{F+T}\right\}$]{\includegraphics[width=0.49\columnwidth]{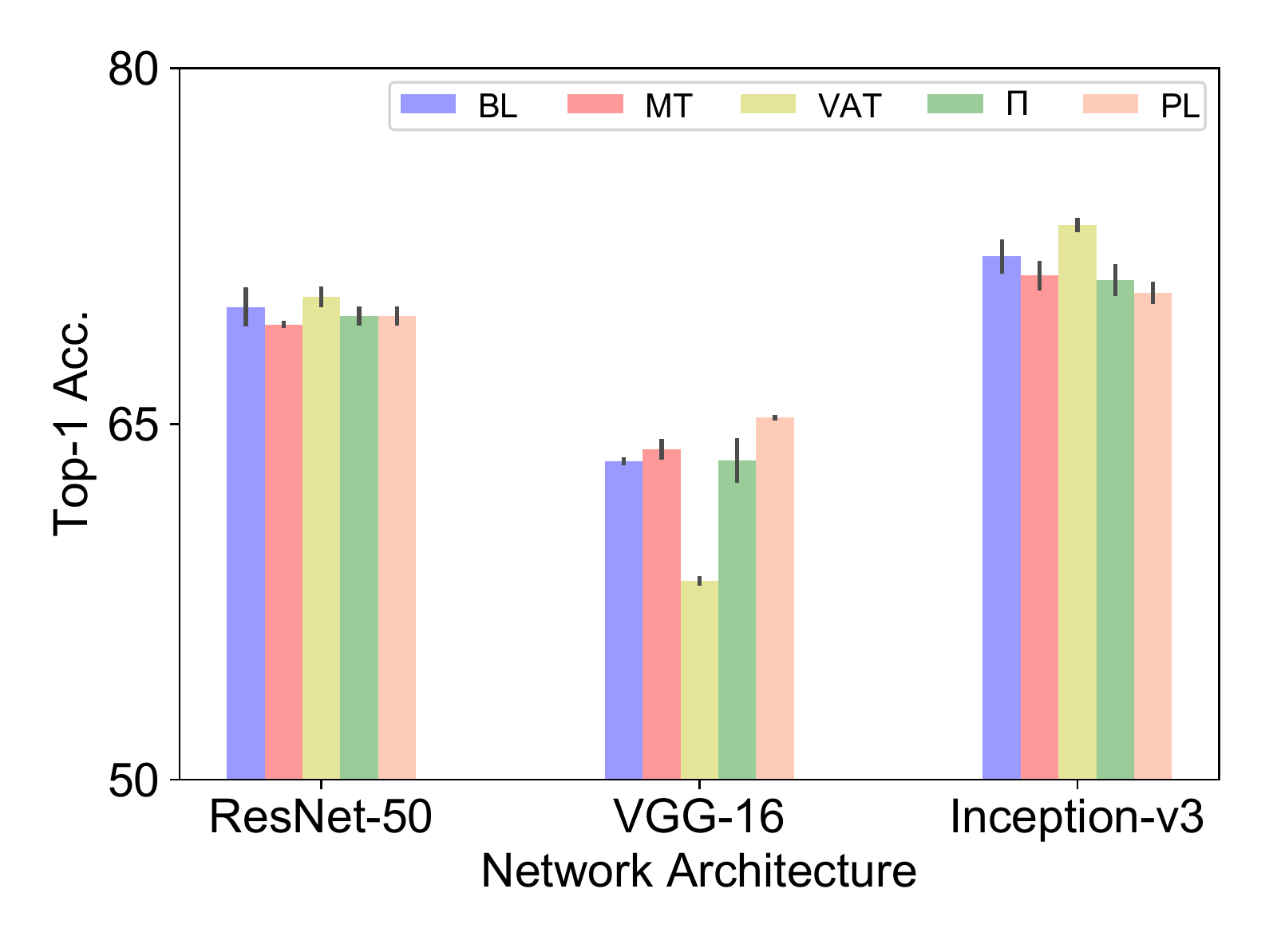}\label{sgd_40_3k_models_indoor_2}}
	\quad
	\subfloat[\scriptsize $\left\{\text{SGD}/\text{40}/\text{1k}/\text{\_}/\text{cub}/\text{F+T}\right\}$]{\includegraphics[width=0.49\columnwidth]{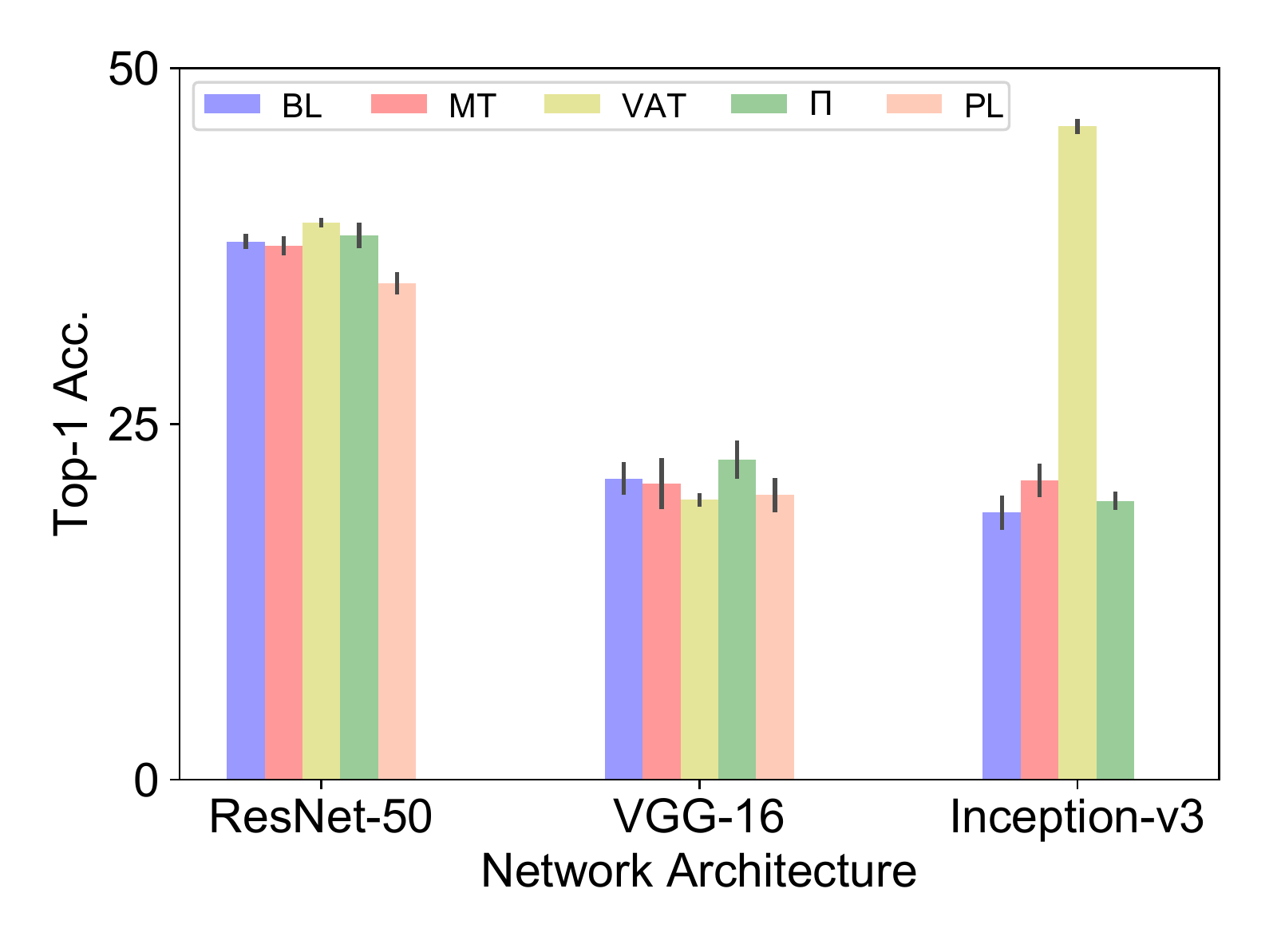}\label{sgd_40_1k_models_cub_2}}
	\subfloat[\scriptsize $\left\{\text{SGD}/\text{40}/\text{4k}/\text{\_}/\text{cub}/\text{F+T}\right\}$]{\includegraphics[width=0.49\columnwidth]{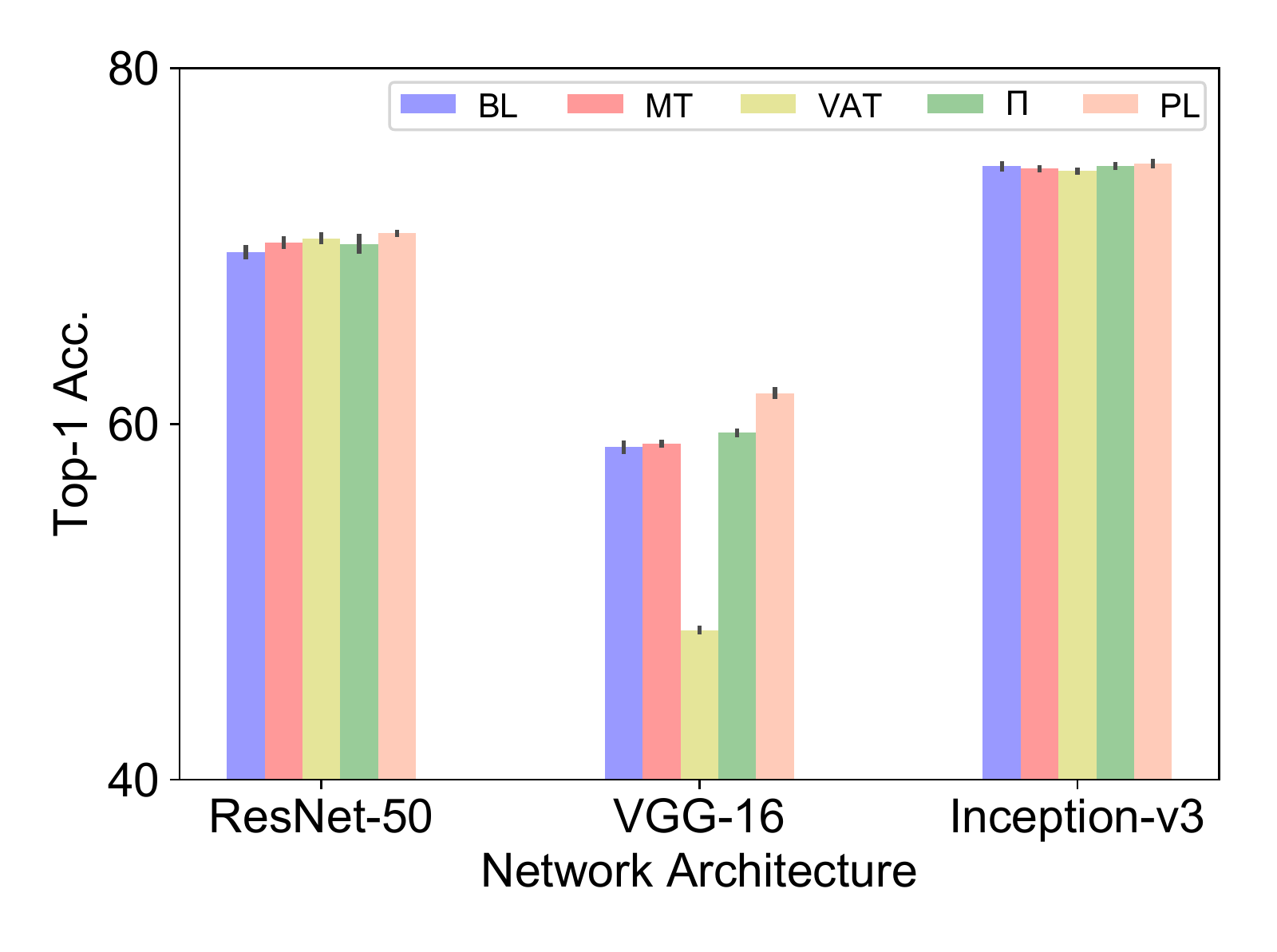}\label{sgd_40_4k_models_cub_2}} \\
	\subfloat[\scriptsize $\left\{\text{SGD}/\text{40}/\text{5360}/\text{\_}/\text{indoor}/\text{F+T}\right\}$]{\includegraphics[width=0.49\columnwidth]{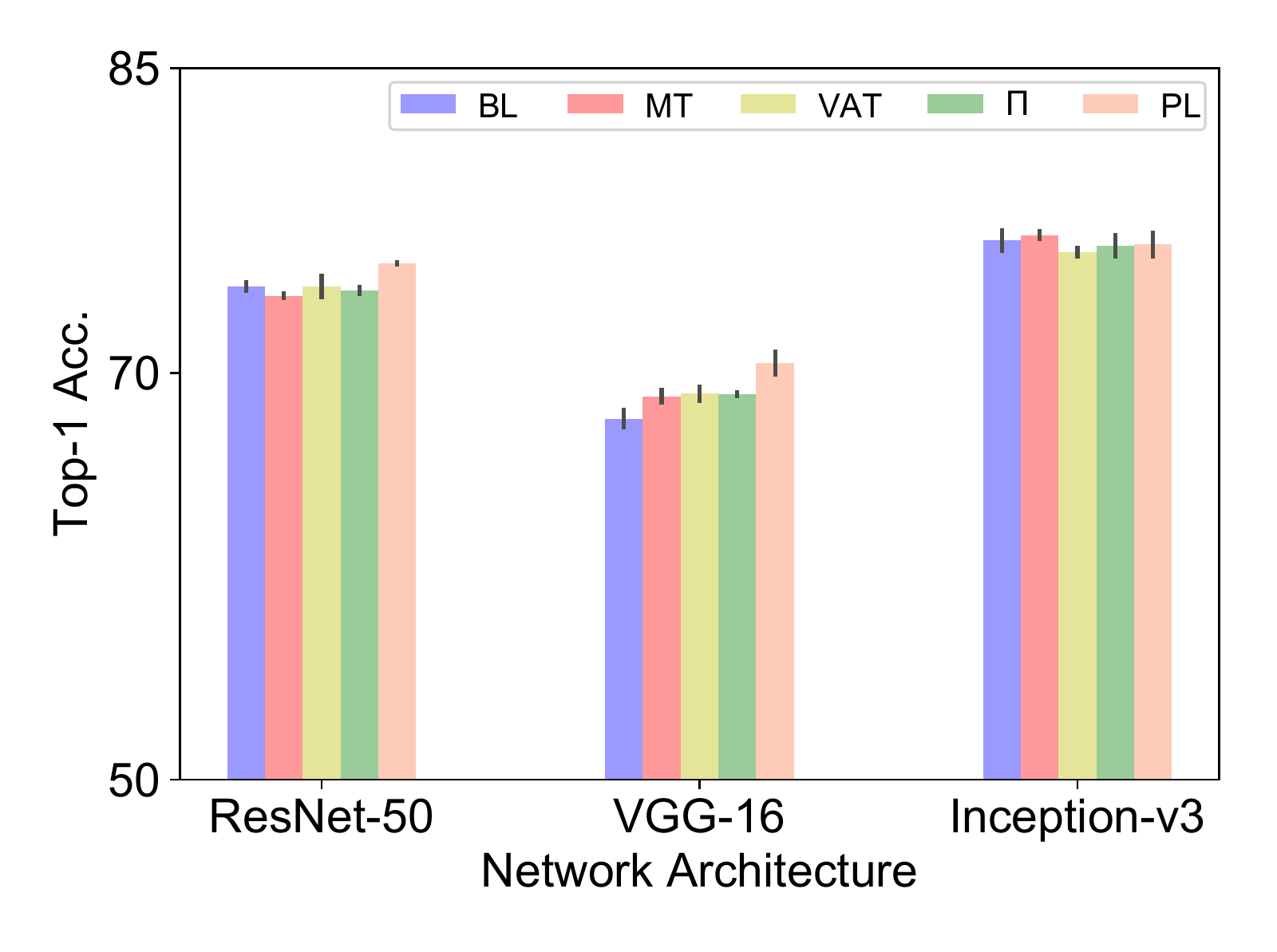}\label{sgd_40_5360_models_indoor_2}}
	\qquad
	\subfloat[\scriptsize $\left\{\text{SGD}/\text{40}/\text{5994}/\text{\_}/\text{cub}/\text{F+T}\right\}$]{\includegraphics[width=0.49\columnwidth]{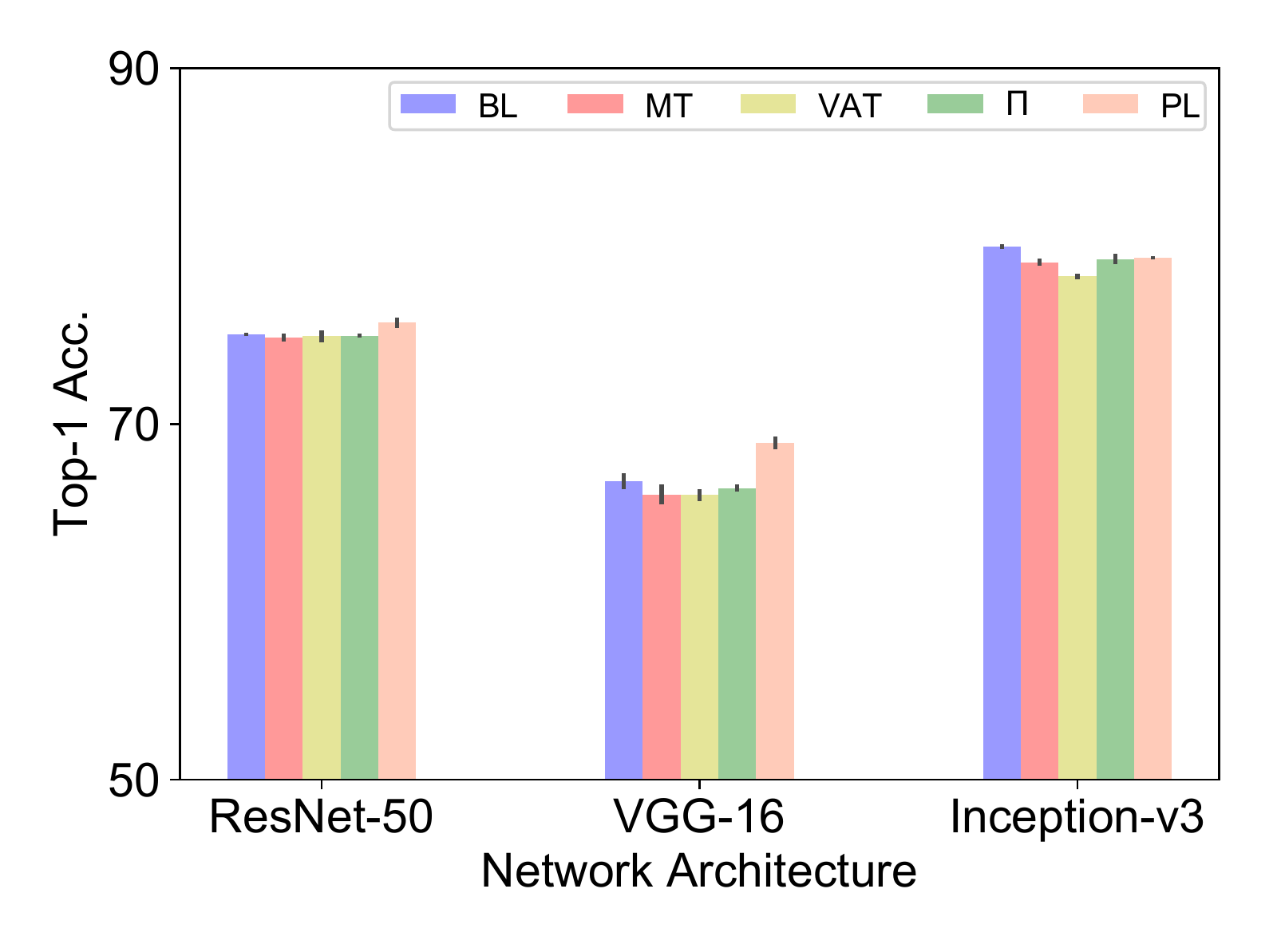}\label{sgd_40_5994_models_cub_2}}
	\qquad
	\subfloat[\scriptsize $\left\{\text{SGD}/\text{40}/\text{2k}/\text{\_}/\text{mura}/\text{F+T}\right\}$]{\includegraphics[width=0.49\columnwidth]{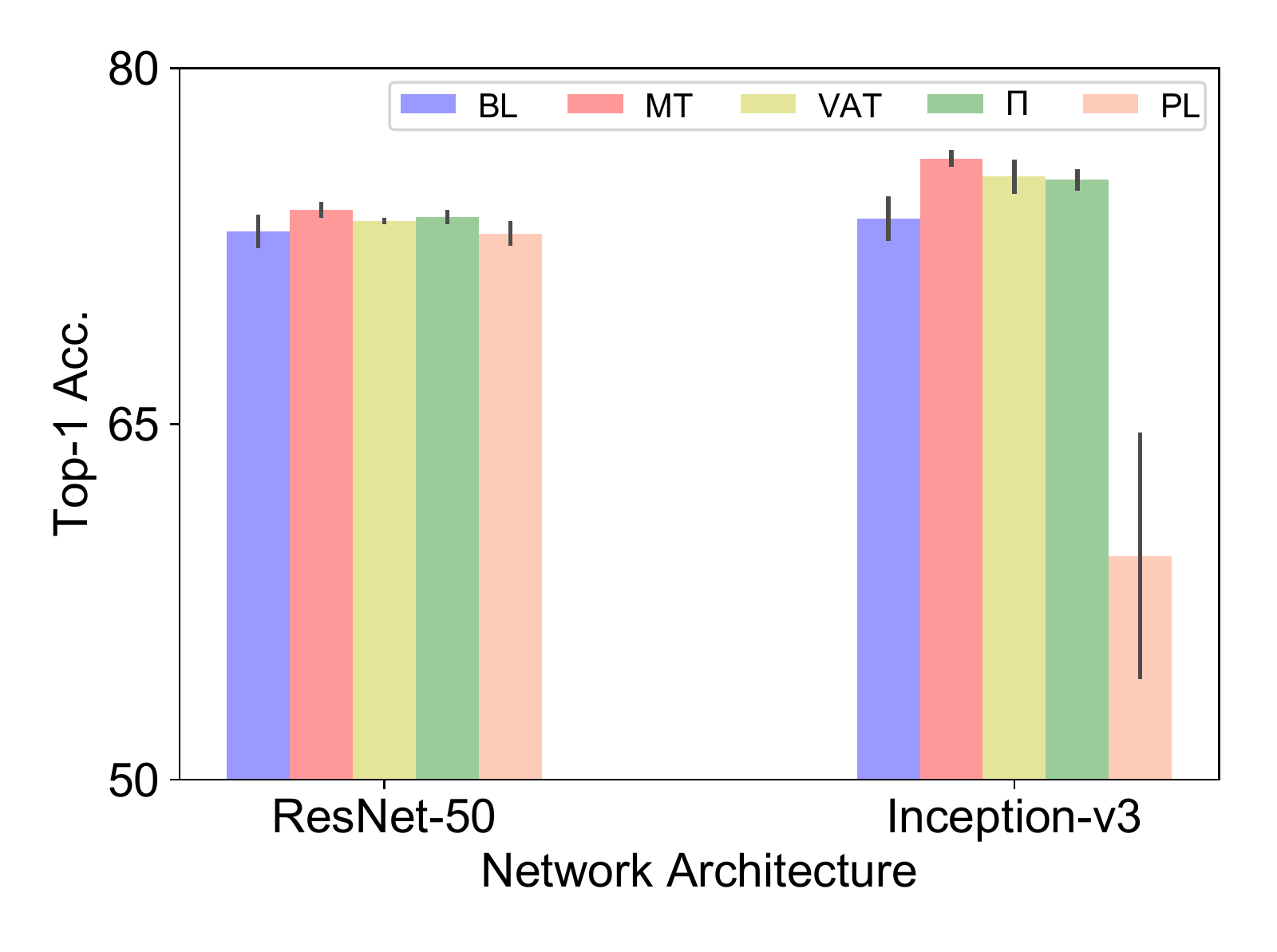}\label{sgd_40_2000_models_mura_2}}
	\caption{SSL algorithms are evaluated on different pre-trained models and datasets. We also increase the amount of labeled data to ensure the reliability of these experiments. Note that all experiments are performed with using SGD as the default optimizer.}
	\label{exp_models}
\end{figure*}
Figure~\ref{exp_models} shows our experiments on different network architectures across three datasets. We choose ResNet-50, VGG-16 and Inception-v3 pre-trained on ImageNet as our basis. As before, for each of these architectures, we fine-tune on Indoor67, CUB200 and MURA, with and without SSL methods. For consistency with those experiments above, we also studied their performance over varying amounts of labeled data.

Figure~\ref{sgd_40_1k_models_indoor_2} and~\ref{sgd_40_1k_models_cub_2} show that \emph{fine-tuning from Inception-v3 trained on ImageNet to other SSL datasets with little labeled data using VAT works very well}. Though Inception-v3 has inferior transfer ability when there is no unlabeled data present, or with other SSL techniques, VAT still helps it get the best performance, defeating ResNet-50 and VGG-16. This points to using Inception series models as a good starting point for fine-tuning with unlabeled data and VAT.

When we increase the amount of labeled data, the superiority of Inception-v3 becomes apparent. If the ratio of labeled images to unlabeled images is roughly two-thirds (cf. Figure~\ref{sgd_40_3k_models_indoor_2} and~\ref{sgd_40_4k_models_cub_2}) or even 100\% (cf. Figure~\ref{sgd_40_5360_models_indoor_2} and~\ref{sgd_40_5994_models_cub_2}), Inception-v3 win the first place among three models. An important fact is that \emph{when you have enough labeled images, better pre-trained models seem to counteract the influence of SSL methods}. When we talk about VGG-16, it is easy to summarize that SSL methods help a lot. In Figure~\ref{sgd_40_3k_models_indoor_2}-~\ref{sgd_40_5994_models_cub_2}, SSL achieves better results than fully-supervised baseline. If we come to ResNet-50, this improvement shrinks but still exists.
Finally, when we look at Inception-v3 in Figure~\ref{sgd_40_3k_models_indoor_2},~\ref{sgd_40_4k_models_cub_2},~\ref{sgd_40_5360_models_indoor_2} and \ref{sgd_40_5994_models_cub_2}, the improvement is gone. Furthermore, fully-supervised method nearly beats all SSL approaches. We are not certain if this means SSL is useless on a good pre-trained model (Inception-v3 scores the best on ImageNet), but this phenomenon can be a good clue to evaluate SSL methods on other well-trained models, such as ResNeXt~\cite{resnext2017} and SE-Net~\cite{senet2018}. However, it is undeniable that SSL performs well when there is a small number of labeled images, or when the domain gap between source and target training is high.

From Figure~\ref{sgd_40_2000_models_mura_2}, we once again find that SSL algorithms work the best under a domain gap. Even on Inception-v3, Mean Teacher still achieves a much better result than fully-supervised method which hints that SSL can be expected in the area of medical image processing.

\begin{figure}[!htp]
	\setlength{\belowcaptionskip}{-0.3cm}
	\scriptsize
	\centering
	\subfloat[\scriptsize $\left\{\text{SGD}/\text{25}/\text{500}/\text{res50}/\text{indoor}/\text{\_}\right\}$]{\includegraphics[width=0.48\columnwidth]{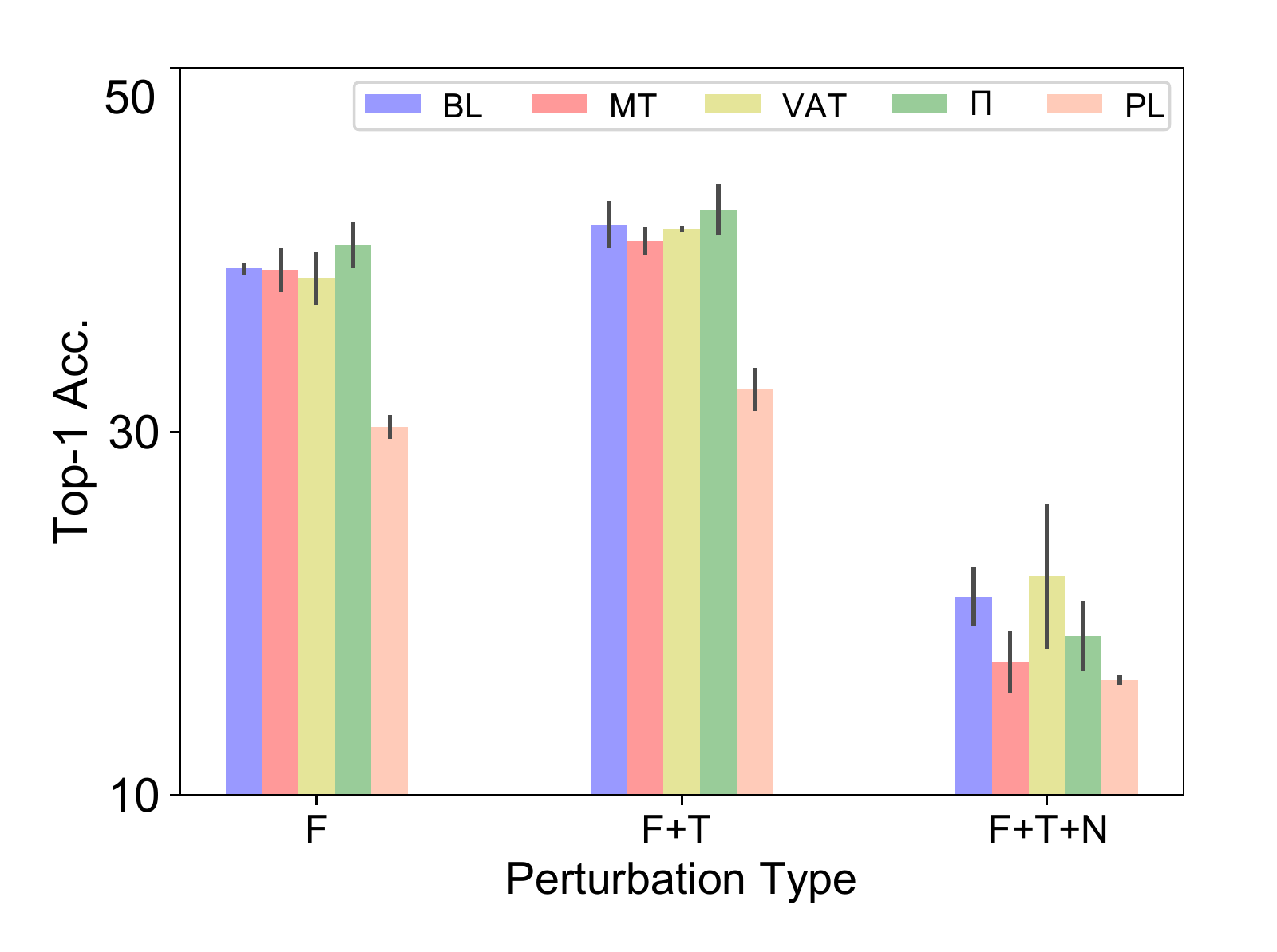}\label{sgd_25_500_res50_indoor_perturb}}
	\quad
	\subfloat[\scriptsize $\left\{\text{SGD}/\text{\_}/\text{2k}/\text{res50}/\text{cub}/\text{F+T}\right\}$]{\includegraphics[width=0.48\columnwidth]{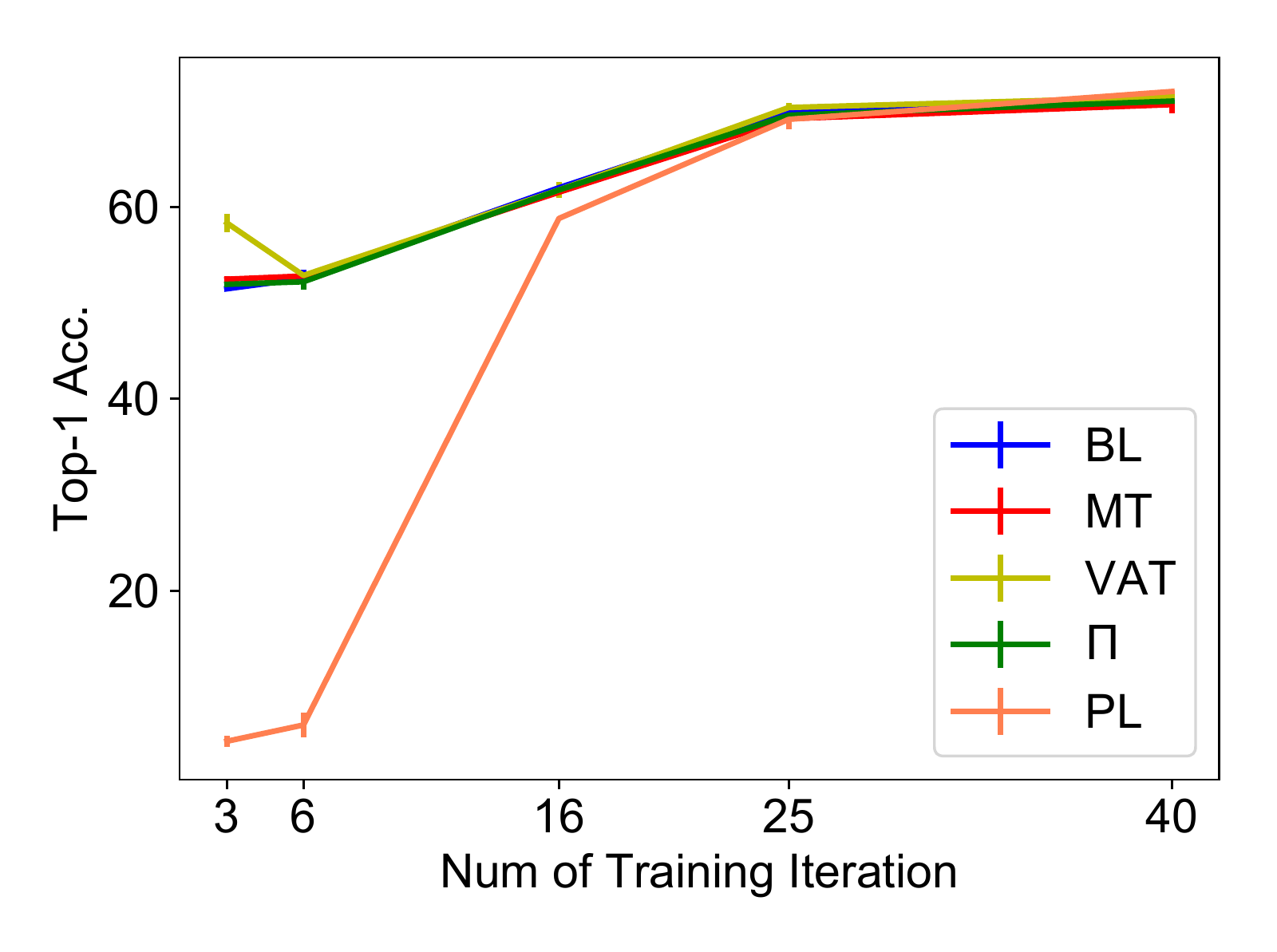}\label{sgd_step_2k_res50_indoor_1}}
	\caption{Experiments on different types of perturbation and varying numbers of training iteration. All experiments use SGD as the default optimizer.}
	\label{perturb_and_iter}
\end{figure}

\begin{figure*}
	\centering
	\subfloat[$\left\{\text{adam}/\text{40}/\text{1k}/\text{res50}/\text{indoor}/\text{F+T}\right\}$]{\includegraphics[width=0.8\columnwidth]{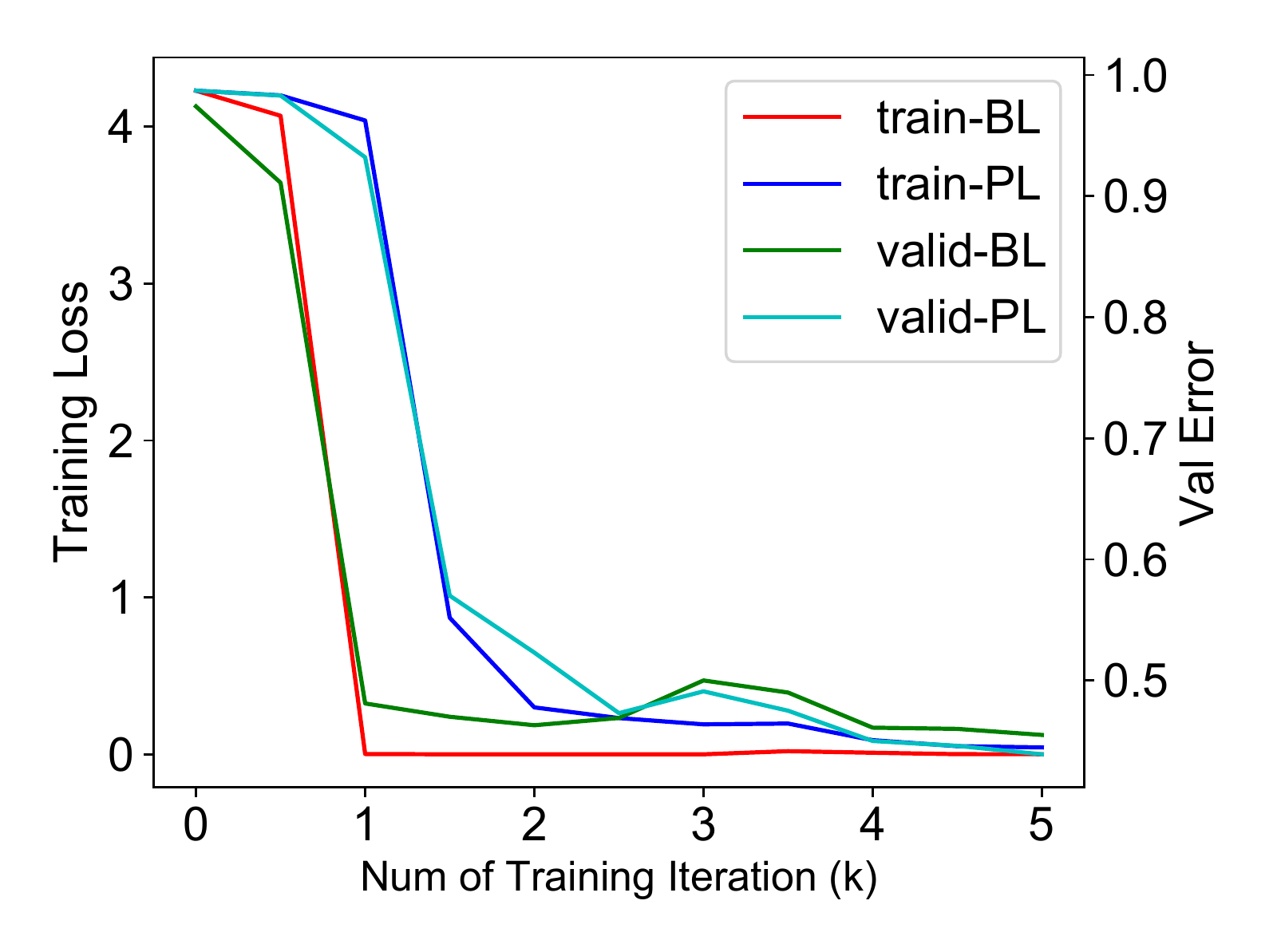}\label{indoor_PL_adam}}
	\quad
	\subfloat[$\left\{\text{adam}/\text{40}/\text{1k}/\text{res50}/\text{cub}/\text{F+T}\right\}$]{\includegraphics[width=0.8\columnwidth]{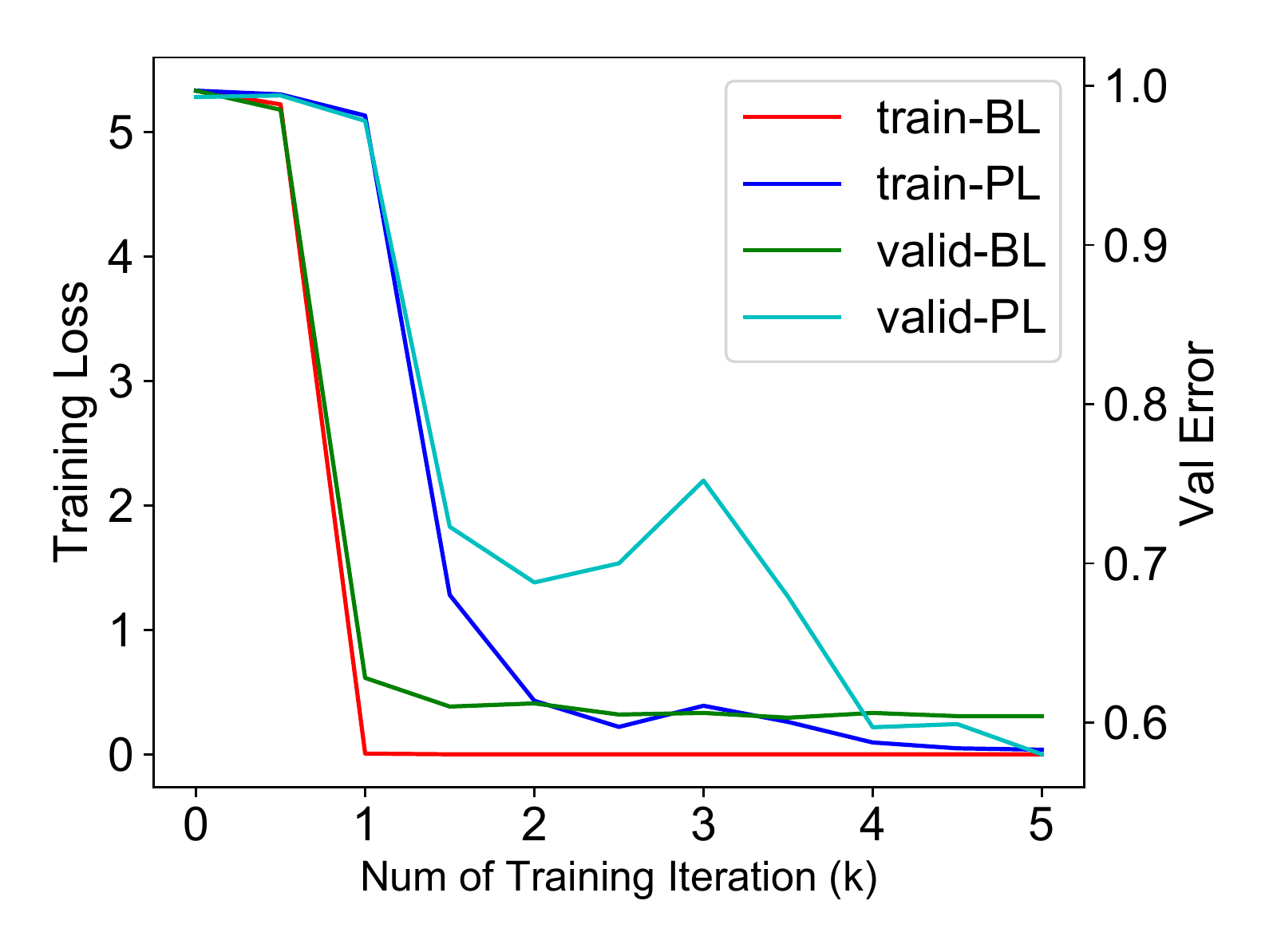}\label{cub_PL_adam}}
	\caption{We report the training and validation details of BL and PL. The horizontal axis represents the number of training iteration and the unit is $k$ (thousand). The training loss means the loss on labeled images.}
	\label{PL_adam}
\end{figure*}

\subsection{Perturbation and Increased Training Iteration}
As shown in the left figure in Figure~\ref{perturb_and_iter}, combining horizontal flip and random translation achieves the highest accuracy among three types of perturbation. On the other hand, simply adding augmentation like noise degenerates the performance. These are the reason why \texttt{F+T} is the first choice in most of our experiments. We also discover that all SSL methods \emph{agree on which perturbation leads to better performance}.

Figure~\ref{sgd_step_2k_res50_indoor_1} suggests \emph{increasing the number of training iteration will close the gap} among different SSL methods. When models are trained for less than 10 epochs, both VAT and Pseudo-Label have unstable performance. However, if we increase the number of training steps, their performances gradually converge.

\section{Conclusion}
In this work, we identify the relationship between SSL and transfer learning. We provide a view that most SSL algorithms can be regarded as a type of smooth regularization, which hints that recent SSL methods may help improve fully-supervised models. To verify the effectiveness of SSL on pre-trained models, we perform detailed experiments under various conditions. The experiments provide several meaningful observations including the effectiveness of SSL on fully supervised methods. Different SSL algorithms usually display different performance under various conditions. However, we find that SSL algorithms are quite useful on medical images. This phenomenon suggests that domain gap might be the place where SSL and transfer learning combined can yield the most improvements.

\appendix
\section*{Appendix}
\section{The regularization effect of Pseudo-Label}
\label{pl}
In order to understand the regularization effect of Pseudo-Label, we show the performance of both training and validation process in Figure~\ref{PL_adam}. The experiments are conducted on Indoor67 and CUB200. We split the original training set (5360 for Indoor67, 5994 for CUB200) into two parts: train and valid and report the training loss and valid errors. The validation set of each dataset contains 1k images. From Figure~\ref{indoor_PL_adam}, we can see that fully-supervised model (BL) achieves lower training loss when compared with Pseudo-Label (PL). In contrast, PL scores lower validation error after 3$k$ iterations. If we focus on CUB200 (Figure~\ref{cub_PL_adam}), it is obvious that BL converges faster and better on the training set while PL gets a lower validation error. These phenomenons hint that \emph{PL may serve as a regularizer and helps prevent overfitting when there is limited labeled data}.

\section{Hyperparameters of VAT}
\label{vat_hyparam}
\begin{figure}[h]
	\centering
	{\includegraphics[width=0.8\columnwidth]{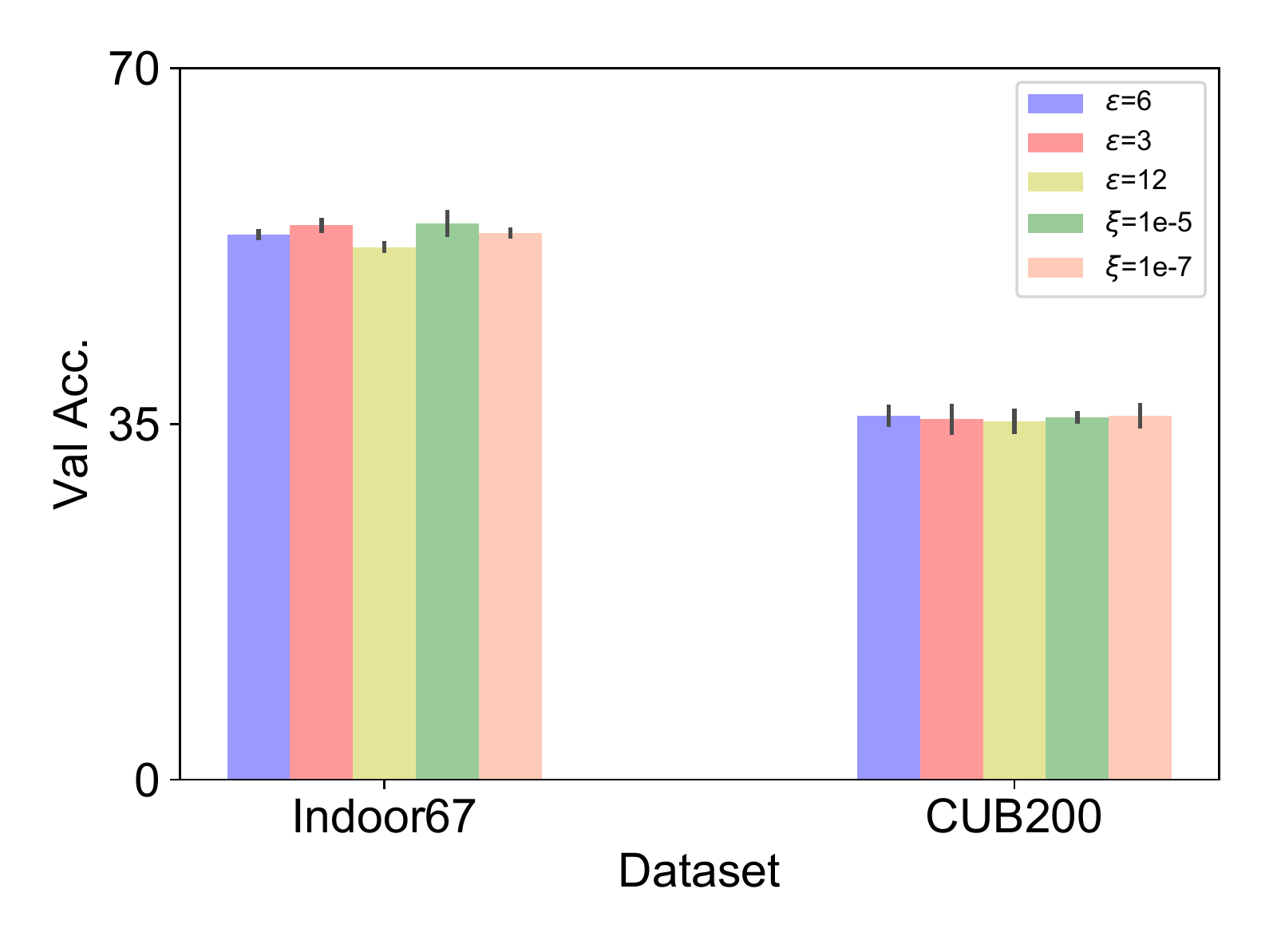}}
	\caption{We compare the performance of VAT using different hyperparameters.}
	\label{vat_studies}
\end{figure}
Since VAT has two hyperparameters: $\epsilon$ and $\xi$, we try to study the influence of them on SSL performance. We report the validation performance of VAT on the same valid set as mentioned in Section~\ref{pl}. The default values of $\epsilon$ and $\xi$ are 6 and 1e-6, respectively. From Figure~\ref{vat_studies}, we find that simply decreasing $\epsilon$ (red bar) or increasing $\xi$ (green bar) would have positive effects on Indoor67. However, when we come to CUB200, the default values achieve comparable results. To keep pace with previous studies, we decide to maintain the default choices of $\epsilon$ (6) and $\xi$ (1e-6) in the paper.

\section{Learning rates on Inception-v3}
\begin{figure}[t]
	\setlength{\belowcaptionskip}{-0.3cm}
	\centering
	{\includegraphics[width=0.8\columnwidth]{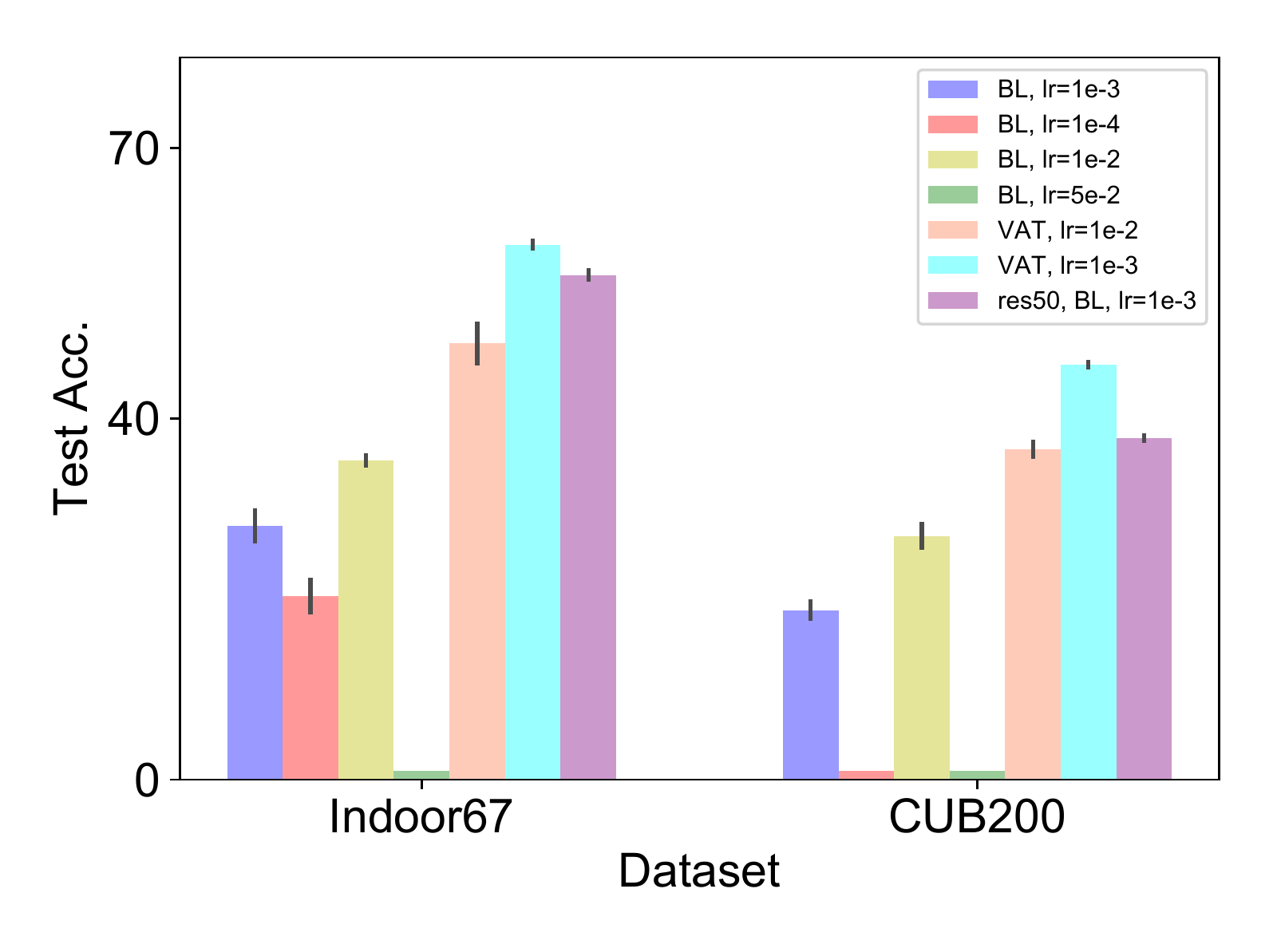}}
	\caption{Influence of different learning rates on Inception-v3.}
	\label{lr_inception}
\end{figure}
It is easy to find that Inception-v3 has inferior performance when having limited labeled data (figure 5 (a) and (c) in the paper). We also conducted complementary ablative studies on this issue and find that learning rate plays an important role. Note that models are evaluated on the test set which is different from the validation set in Section~\ref{pl} and \ref{vat_hyparam}. We report the experimental results in Figure~\ref{lr_inception}. All training details follow the settings in figure 5 (a) and (c) (please refer to the paper for more details). From Figure~\ref{lr_inception}, we can see that inappropriate choices of learning rates may heavily influence the performance of Inception-v3, such as lr=5e-2 (red bar) has bad results on both Indoor67 and CUB200. Also, increasing the learning rate to 1e-2 will enhance the model accuracy (comparing blue with yellow) but deteriorate VAT (comparing coral with aqua). 

It is worth noting that there are several reasons that we report the results of training Inception-v3 using lr=1e-3 in figure 5 (cf. our paper). Firstly, fully-supervised Inception-v3 with lr=1e-2 (yellow bar) still cannot defeat res50 (bar in purple). Next, Inception-v3 achieves satisfying results using lr=1e-3 with increased labeled images (refer to figure 5 (b) and (d-g)). Considering the experimental consistency with ResNet-50 and VGG-16, we decide to perform experiments using lr=1e-3.

{\small
	\bibliographystyle{ieee}
	\bibliography{egbib}
}

\end{document}